\documentclass[10pt,a4paper,logo]{qwenapplication}
\usepackage[numbers]{natbib}

\usepackage{amsmath,amsfonts,bm}

\def\eqref#1{equation~\ref{#1}}

\def\1{\bm{1}}

\DeclareMathAlphabet{\mathsfit}{\encodingdefault}{\sfdefault}{m}{sl}
\SetMathAlphabet{\mathsfit}{bold}{\encodingdefault}{\sfdefault}{bx}{n}

\usepackage{amsmath}
\usepackage{amssymb}
\usepackage{array}
\usepackage{booktabs}
\usepackage{colortbl}
\usepackage{flafter}
\usepackage{float}
\usepackage{graphicx}
\usepackage{hyperref}
\usepackage{listings}
\usepackage{multirow}
\usepackage{placeins}
\usepackage{tcolorbox}
\usepackage{url}
\usepackage{xcolor}
\tcbuselibrary{breakable,skins}

\newtcolorbox{judgepromptbox}[1][]{
  enhanced,breakable,
  colback=black!3,colframe=black!65,
  boxrule=0.5pt,arc=2pt,
  left=8pt,right=8pt,top=10pt,bottom=8pt,
  before skip=14pt,after skip=14pt,
  fonttitle=\bfseries,coltitle=white,
  title={Judge Prompt},
  title after break={Judge Prompt (continued)},
  extras title after break={halign title=center},
  attach boxed title to top center={yshift=-2mm},
  boxed title style={colback=black!70,colframe=black!70,arc=1pt},
  #1
}
\newtcolorbox{benchmarkexamplebox}[1][]{
  enhanced,breakable,
  colback=white,colframe=black!65,
  boxrule=0.5pt,arc=2pt,
  left=8pt,right=8pt,top=8pt,bottom=8pt,
  before skip=12pt,after skip=12pt,
  fonttitle=\bfseries,coltitle=white,
  colbacktitle=black!70,
  halign title=center,
  title={An example of \dataset{}},
  title after break={An example of \dataset{} (continued)},
  #1
}
\lstdefinestyle{judgejson}{
  basicstyle=\ttfamily\normalsize,
  columns=fullflexible,keepspaces=true,
  breaklines=true,breakatwhitespace=false,
  showstringspaces=false,
  aboveskip=6pt,belowskip=6pt
}

\newcommand{\dataset}{\textsc{MemCalib}}
\newcommand{\method}{\textsc{MemCalib-RL}}
\newcommand{\ignorelevel}{\textsc{Ignore}}
\newcommand{\boundlevel}{\textsc{Bound}}
\newcommand{\controllevel}{\textsc{Control}}
\newcommand{\levelideal}{\ell}
\newcommand{\levelhat}{\hat{\ell}}
\newcommand{\smos}{\textsc{sMOS}}
\newcommand{\smus}{\textsc{sMUS}}
\newcommand{\aor}{\textsc{AOR}}
\newcommand{\aur}{\textsc{AUR}}

\title{\dataset: Benchmarking and Optimizing\\
Memory Use in LLM Agents}

\author[1,2,*]{Ruike Cao}
\author[2,3,*]{Fanyu Zhao}
\author[2,\textdagger]{Fugen Yao}
\author[2]{Liang Dong}
\author[2]{Jian Xu}
\author[3]{Yifei Zhao}
\author[3]{Han Zhang}
\author[1,\textdagger]{Li Xiao}

\affil[1]{University of Science and Technology of China}
\affil[2]{Qwen Business Unit of Alibaba}
\affil[3]{Fudan University}
\authornote{\textsuperscript{*}Equal contribution.
\textsuperscript{\textdagger}Corresponding authors.}

\begin{abstract}
The effectiveness of agent memory ultimately depends on whether the underlying
LLM gives each memory in context an appropriate degree of influence over its
response. Yet this capability has remained largely overlooked. To assess this capability, we
introduce \dataset{}, a
benchmark grounded in realistic memory-system scenarios for evaluating
memory use and advancing optimization algorithms. Results on the \dataset{} test set reveal that
frontier open- and closed-source models struggle to use memory appropriately.
They frequently over-use or under-use memory rather than matching each
proposition's actual use to its target level, leading to biased, low-quality
responses.
Experiments with common post-training algorithms, including group relative
policy optimization and on-policy self-distillation,
further reveal a clear directional skew: trained models improve in one
direction while deteriorating in the other. We therefore propose
\method{}, an ordered bidirectional counterfactual credit-assignment
algorithm that separates over- and under-use signals and localizes their credit
to response tokens through exact atom ablation. Results across model families
and scales (\mbox{Qwen3-8B}, Ministral-3-8B-Instruct, and Qwen3.5-35B-A3B) show that
\method{} achieves the best overall performance while better balancing over-use
and under-use, with gains generalizing beyond \dataset{} in external benchmark
evaluation. Further experiments support its design choices and
robustness and provide insight into its training dynamics.\footnote{Code and data: \url{https://github.com/Quark-Medical/memcalib}}
\end{abstract}

\begin{document}

\maketitle

\begingroup
\renewcommand{\thefootnote}{\textdagger}
\footnotetext[0]{Corresponding authors: Fugen Yao
(\href{mailto:fugen.yfg@alibaba-inc.com}{\nolinkurl{fugen.yfg@alibaba-inc.com}})
and Li Xiao (\href{mailto:xiaoli11@ustc.edu.cn}{\nolinkurl{xiaoli11@ustc.edu.cn}}).}
\endgroup

\section{Introduction}
\label{sec:intro}

Recent advances in reasoning, planning, and coding capabilities, together with
the integration of memory and tools, have transformed LLMs from static
conditional generators into adaptive policies that interact with
external environments \citep{yehudai2025agentsurvey}. This transformation has
given rise to LLM-based agents, in which memory serves as a core capability
supporting long-horizon reasoning, continual adaptation, and coherent
interaction with complex environments
\citep{hu2026agentmemorysurvey,luo2026storage}.
Although memory systems may differ in how they form, update, retrieve, and
filter memories, their effectiveness ultimately depends on whether the LLM
gives each memory in context the appropriate level of influence.
Post-retrieval filtering cannot guarantee a noise-free context: some supplied
memories may remain irrelevant or outdated, while even useful memories differ
in the scope of influence they should have. Moreover, real memory systems
often return summaries, profiles, or trajectories in which a single memory
block contains multiple atomic propositions
\citep{zhong2024memorybank,park2023generativeagents,li2026timem}.
Within the same block, some atoms
may be noise, some should provide only local support, and others should control
a core conclusion. The model must therefore make a finer decision than
use or ignore: it must determine, at the atomic level, how each
proposition should shape its response.
\begin{figure}[t]
\centering
\includegraphics[width=\textwidth]{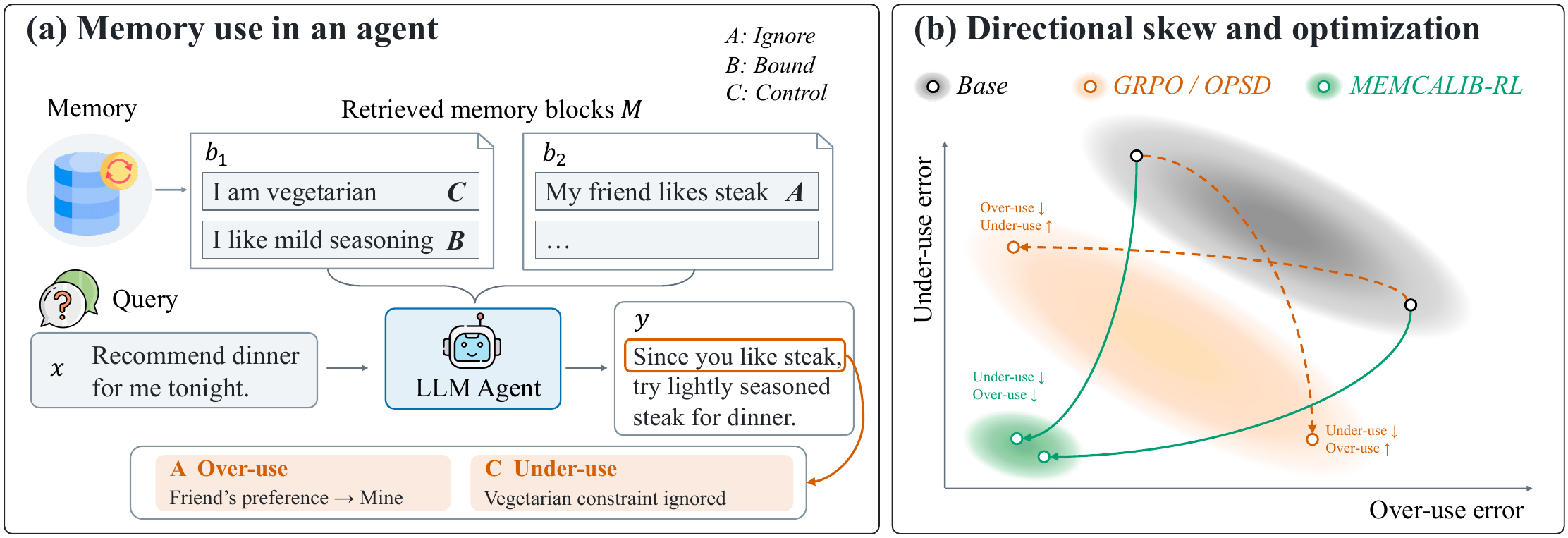}
\caption{Memory use and optimization.
(a) An example of LLM agents over-using irrelevant memory and under-using
relevant constraints.
(b) A schematic comparison of optimization behavior: GRPO and OPSD can reduce
both errors in principle but tend to prioritize one objective in practice due
to coarse credit assignment, while \method{} better balances the two for
better overall performance.}
\label{fig:overview}
\end{figure}
For each atomic
proposition in the supplied memory context, its appropriate level of influence
on the response is determined based on the current task: the proposition should leave
no answer-specific footprint (\ignorelevel{}),
provide bounded local support (\boundlevel{}), or control a material
conclusion (\controllevel{}). Influence that exceeds or falls below the target
level is harmful: over-use can let irrelevant or outdated memories override
current evidence, distort core conclusions, or cause over-personalization
\citep{hu2026opbench,xiang2026memsyco};
under-use can discard preferences, constraints, and experience that should
govern the answer
\citep{feng2026rpeval,yoon2026benchpres,xiong2026impact}. Given that the
effectiveness of agent memory systems ultimately depends on the LLM's
ability to use memory appropriately, a fundamental question remains largely
overlooked:

{\centering
\emph{Are LLMs really capable of using memory
appropriately to shape their responses?}\par}

Recent personalization benchmarks have already revealed limitations in
LLM-based personalized assistants: they may over- or under-use personalized
preference memories, causing them to over-personalize, become sycophantic, or
fail to apply relevant preferences
\citep{hu2026opbench,xiang2026memsyco,feng2026rpeval,yoon2026benchpres}
(see Appendix~\ref{app:related-work} for further discussion).
Taken together, these limitations motivate us to re-examine LLMs' ability to
use memory appropriately. We therefore introduce \dataset{}, a benchmark
spanning health, general assistance, and coding, comprising 15,000 examples:
a 13,500-example training set to facilitate innovation in optimization
algorithms and a disjoint 1,500-example test set for evaluation. \dataset{} captures a
realistic challenge faced by LLMs
in agent systems: given a query and a variable number of natural composite
memory blocks, each of which contains a variable number of atomic
propositions, a model must calibrate each proposition's influence on its
response. For evaluation, an LLM-based judge \citep{zheng2023judging} applies
atom-specific rubrics to classify the model's
actual use of each atom in its response as \ignorelevel{}, \boundlevel{}, or
\controllevel{}.
Comparing the target and actual use levels then quantifies over-use, under-use,
and overall memory-use performance.

Results on the test set reveal widespread mismatches between atoms' actual and
target use levels across frontier open- and closed-source models. Most models
also exhibit clear
\emph{directional skew}: they perform relatively well with respect to one 
error type but worse on the other, indicating a globally
aggressive or conservative memory-use policy rather than calibrating each 
proposition to the current query. Moreover, experiments show that widely
used post-training algorithms, including group relative policy optimization
(GRPO) \citep{shao2024deepseekmath} and on-policy self-distillation (OPSD)
\citep{zhao2026selfdistilled}, also exhibit clear directional skew: the trained
models improve in one direction while deteriorating in the
other, a phenomenon
which we call the \emph{calibration seesaw}.
We therefore propose \method{}, an ordered bidirectional counterfactual
credit-assignment method that decomposes the matrix of target and actual
use levels into nine reward channels and then uses exact atom ablation and
target-aware counterfactual likelihood differences to localize each channel's
credit to response tokens. Across model families and scales (Qwen3-8B,
Ministral-3-8B-Instruct, and Qwen3.5-35B-A3B), \method{} delivers the best
overall performance while better balancing over-use
and under-use.
Figure~\ref{fig:overview} summarizes the memory-use challenge
and the contrasting optimization directions.
Furthermore, these gains extend
beyond \dataset{} to the external RPEval benchmark \citep{feng2026rpeval}.
Ablation and sensitivity analyses support our localization design. Mechanism analyses reveal how credit allocation
evolves during training, while evaluation with an alternative Judge supports the
robustness of the overall method comparison.
Our contributions are:
\begin{itemize}
    \item We identify a long-overlooked problem: LLMs
    often fail to use memory appropriately to shape their responses and exhibit
    clear directional skew between over-use and under-use.
    \item We introduce \dataset{}, a benchmark that evaluates whether models
    match each proposition's influence to its target level within natural
    composite memory blocks across health, general assistance, and coding.
    \item We propose \method{}, an ordered bidirectional counterfactual
    credit-assignment algorithm. Extensive experiments across model families,
    scales, and benchmarks demonstrate consistent gains and a better balance
    between over-use and under-use.
\end{itemize}

\section{Benchmarking Memory Use in LLMs}
\label{sec:benchmark}

\subsection{Problem formulation}
Consider an LLM-based agent that receives a user query together with memory
blocks retrieved from its memory system. The LLM must determine how each
block's constituent propositions should shape its response. Let \(x\) denote
the query, \(M=\{b_n\}_{n=1}^{N}\) the supplied memory blocks, and \(\pi_\theta\)
the LLM policy with parameters \(\theta\). The model generates
\(y\sim\pi_\theta(\cdot\mid x,M)\). Each block \(b_n\)
comprises \(K_n\) atomic propositions,
\(b_n=\{m_{n,k}\}_{k=1}^{K_n}\).
For the current query \(x\), each atomic proposition \(m_{n,k}\) has an ideal
use level \(\levelideal_{n,k}\in
\{\ignorelevel{},\boundlevel{},\controllevel{}\}\), where
\ignorelevel{} leaves no answer-specific footprint, \boundlevel{} provides
bounded local support, and \controllevel{} determines a material conclusion,
constraint, or recommendation. An LLM that uses memory appropriately should
let each atom influence its response at that atom's ideal level.

\subsection{Benchmark Construction}
\label{sec:construction}

We therefore construct \dataset{} from eight public question-answering datasets
spanning health, general assistance, and coding to evaluate whether LLMs can use
memory appropriately across a broad range of scenarios. Our core construction
strategy is to separate each source sample into a current query and memory
atoms containing information not included in the query, and annotate the
extracted atoms with ideal use levels. The source context generally supports
the current request, so most extracted atoms are labeled \boundlevel{} or
\controllevel{}; we therefore add controlled \ignorelevel{} distractor atoms.
We then generate atom-specific rubrics for all atoms, specifying response-text
criteria for assessing their actual use levels. The atoms are then assembled
into composite blocks and rewritten as fluent memory text while preserving
every proposition.
To ensure data quality, we use a six-stage LLM-based construction pipeline
with deterministic checks and two independent LLM-based semantic reviews,
which is iteratively refined based on human review of freshly constructed pilot
batches and then frozen for full-scale construction
(see Appendix~\ref{app:construction-process} for details).
Using this pipeline, we construct 15,000 \dataset{} examples spanning health,
general assistance, and coding, with a stratified partition yielding 13,500
training and 1,500 disjoint test examples. Detailed
benchmark statistics are provided in Appendix~\ref{app:benchmark-composition}.

\begin{table*}[t]
\caption{Evaluation results of representative models on the \dataset{} test
set (mean \(\pm\) standard deviation over three seeds). Red and blue denote the
best and second-best result in each column.}
\label{tab:model-survey}
\begin{center}
\setlength{\tabcolsep}{2.15pt}
\begin{tabular}{@{}lcccccc@{}}
\toprule
Model & SCS \(\uparrow\) & Exact \(\uparrow\) & \smos{} \(\downarrow\) &
\smus{} \(\downarrow\) & \aor{} \(\downarrow\) & \aur{} \(\downarrow\)\\
\midrule
GPT-5.6-SOL & \(\textcolor{red}{46.25{\pm}0.60}\) & \(\textcolor{red}{28.40{\pm}0.85}\) & \(\textcolor{blue}{38.45{\pm}0.15}\) & \(22.92{\pm}0.40\) & \(\textcolor{blue}{53.33{\pm}0.45}\) & \(35.82{\pm}0.51\)\\
Claude Sonnet 4.6 & \(\textcolor{blue}{36.44{\pm}0.34}\) & \(19.09{\pm}0.33\) & \(48.89{\pm}0.62\) & \(24.77{\pm}0.37\) & \(65.40{\pm}0.77\) & \(37.93{\pm}0.38\)\\
Kimi-K2.6 & \(35.54{\pm}0.18\) & \(\textcolor{blue}{20.00{\pm}0.28}\) & \(53.83{\pm}0.19\) & \(\textcolor{blue}{19.94{\pm}0.60}\) & \(68.67{\pm}0.63\) & \(\textcolor{blue}{31.56{\pm}0.98}\)\\
Gemini 3.5 Flash & \(34.96{\pm}0.53\) & \(17.96{\pm}0.58\) & \(51.65{\pm}0.45\) & \(24.73{\pm}0.14\) & \(68.27{\pm}0.38\) & \(37.69{\pm}0.25\)\\
GLM-5.2 & \(34.40{\pm}0.45\) & \(18.20{\pm}0.61\) & \(52.13{\pm}0.42\) & \(23.42{\pm}0.31\) & \(67.53{\pm}0.88\) & \(35.73{\pm}0.71\)\\
Qwen3.8-Max & \(34.22{\pm}0.52\) & \(17.80{\pm}0.48\) & \(50.11{\pm}0.33\) & \(27.46{\pm}0.32\) & \(66.09{\pm}0.51\) & \(41.16{\pm}0.21\)\\
DeepSeek-V4-Flash & \(33.72{\pm}0.76\) & \(16.96{\pm}0.93\) & \(55.64{\pm}1.08\) & \(20.86{\pm}0.59\) & \(73.31{\pm}1.01\) & \(32.64{\pm}0.76\)\\
Qwen3-8B & \(31.17{\pm}0.21\) & \(15.29{\pm}0.23\) & \(\textcolor{red}{37.19{\pm}0.62}\) & \(45.66{\pm}0.36\) & \(\textcolor{red}{50.24{\pm}0.87}\) & \(61.60{\pm}0.24\)\\
Qwen3.5-35B-A3B & \(26.54{\pm}0.28\) & \(12.24{\pm}0.25\) & \(64.76{\pm}0.48\) & \(\textcolor{red}{19.93{\pm}0.32}\) & \(80.16{\pm}0.83\) & \(\textcolor{red}{31.11{\pm}0.60}\)\\
\bottomrule
\end{tabular}
\end{center}
\end{table*}

\subsection{Evaluation protocol and metrics}
\label{sec:evaluation}

We adopt rubric-guided LLM-as-a-Judge evaluation
\citep{zheng2023judging,kim2024prometheus2,li2025generation} to assess the
actual use level \(\hat{\ell}_{n,k}\) of each atom \(m_{n,k}\) in response
\(y\) as \ignorelevel{}, \boundlevel{}, or \controllevel{}.
The judge prompt is provided in Appendix~\ref{box:judge-prompt}.
To quantify deviations from the ideal levels, we define
\(\operatorname{rank}\) to map \ignorelevel{}, \boundlevel{}, and
\controllevel{} to \(0,1,2\), respectively. We then define the response-level
over-use and under-use totals by summing how far each atom's actual use level
exceeds or falls below its ideal level, respectively:
\[
O=\sum_{n,k}\max\!\left(
\operatorname{rank}(\hat{\ell}_{n,k})-
\operatorname{rank}(\ell_{n,k}),0\right),
\quad
U=\sum_{n,k}\max\!\left(
\operatorname{rank}(\ell_{n,k})-
\operatorname{rank}(\hat{\ell}_{n,k}),0\right).
\]

Let \(O_j\) and \(U_j\) denote the over-use and under-use totals for the
\(j\)-th response in evaluation set \(\mathcal E\), respectively.
With a geometric decay factor \(\rho=0.5\) (each exponential term halves per
unit increase in its exponent), we define
sample-level Memory Overuse Severity (\smos{}) and sample-level Memory
Underuse Severity (\smus{}) as:
\[
\smos=\frac{1}{|\mathcal E|}\sum_{j=1}^{|\mathcal E|}(1-\rho^{O_j}),
\qquad
\smus=\frac{1}{|\mathcal E|}\sum_{j=1}^{|\mathcal E|}(1-\rho^{U_j}).
\]
We also report the Any-Overuse Rate (\aor{}) and Any-Underuse Rate (\aur{})
as the proportions of responses with at least one over-use or under-use error,
respectively.
To assess overall memory-use performance, we define the Sample Calibration
Score (SCS) and Exact Calibration (Exact) accounting for both over-use and
under-use:
\[
\mathrm{SCS}=\frac{1}{|\mathcal E|}\sum_{j=1}^{|\mathcal E|}\rho^{O_j+U_j},
\qquad
\mathrm{Exact}=\frac{1}{|\mathcal E|}\sum_{j=1}^{|\mathcal E|}
\mathbf{1}[O_j+U_j=0].
\]
Higher SCS and Exact scores, together with lower error severity and rates,
indicate more appropriate memory use. All six metrics are defined on \([0,1]\)
and rescaled to \([0,100]\) for reporting.

\subsection{Memory-Use Performance Across Models}
\label{sec:diagnosis}
Having established atom-specific rubrics and comprehensive evaluation metrics,
we evaluate representative open- and closed-source models on the 1,500-example
\dataset{} test set, using DeepSeek-V4-Pro \citep{deepseekai2026deepseekv4}
to assess each atom's actual use level. We run each model in non-thinking mode
using three independent seeds, with both temperature and top-\(p\) set to 1.

The evaluation results in Table~\ref{tab:model-survey} show that all evaluated
models struggle to use memory appropriately, with low SCS and Exact scores
even among frontier closed-source models.
Moreover, larger models do not necessarily use memory more appropriately:
Qwen3-8B outperforms Qwen3.5-35B-A3B in both SCS and Exact.
Beyond SCS and Exact, the directional metrics reveal a clear skew among the
evaluated LLMs: all except Qwen3-8B over-use memory more severely and more
frequently than they under-use it, whereas Qwen3-8B shows the reverse.
GPT-5.6-SOL achieves the strongest overall performance because its errors are
comparatively low in both directions. This outcome also illustrates that SCS
and Exact are comprehensive metrics that take both over-use and under-use into
account: strength in one direction cannot compensate for substantial errors
in the other. This directional skew further suggests that models rely on a
broad prior over whether memory should be trusted, leaving individual
propositions poorly calibrated to the current query. To verify the reliability
of the use-level judgments underlying these results, we have a human annotator
label the actual use levels of sampled atoms.
The results show 96.7\% agreement between the Judge and human annotations on
150 naturally sampled judgments (95\% Wilson CI: 92.4\%--98.6\%; Cohen's
\(\kappa=0.872\)), with disagreements concentrated at the
\boundlevel{}/\controllevel{} boundary (Appendix~\ref{app:judge-audit}).

\begin{figure*}[t]
\centering
\includegraphics[width=\textwidth]{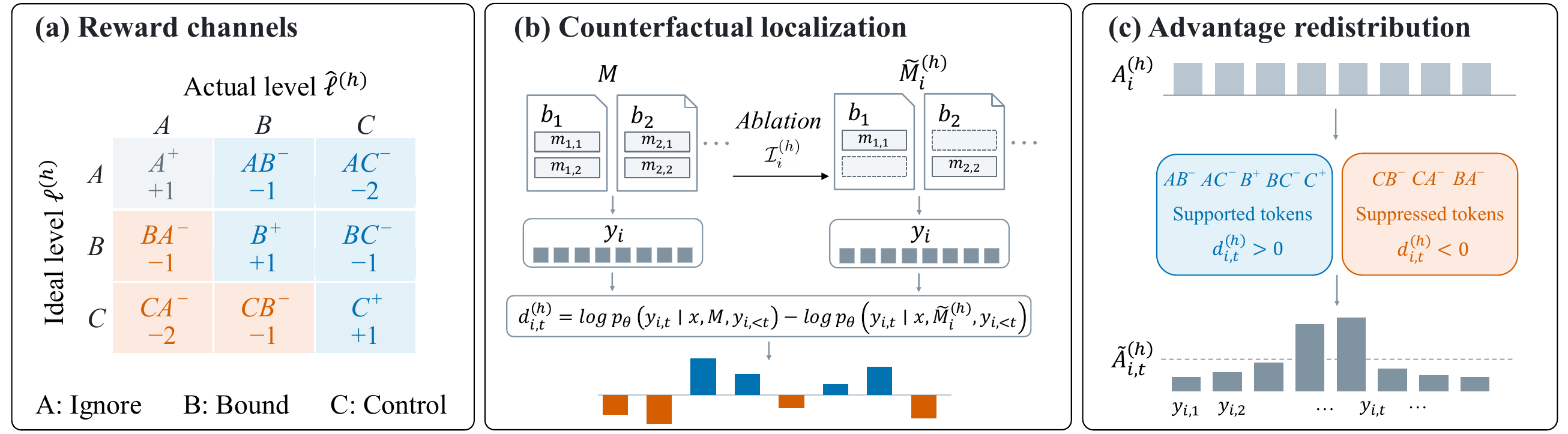}
\caption{Overview of \method{}.
(a) Each atom's ideal and actual use levels determine its assignment to one of nine reward channels.
(b) Exact atom ablation yields token-level log-likelihood differences for the
same response under full and ablated memories.
(c) These signals guide channel-specific advantage redistribution toward
supported or suppressed tokens while preserving each channel's response-level mean.}
\label{fig:method}
\end{figure*}

\section{\method: Ordered Bidirectional Credit Assignment}
\label{sec:method}

A single response can use some memory atoms appropriately while over- or
under-using others, yet standard GRPO collapses these outcomes into a single
advantage shared by every token \citep{shao2024deepseekmath}. In our experiments
(Section~\ref{sec:experiments}), this coarse supervision yields a calibration
seesaw: reducing one error type increases the other. We therefore propose
\method{} (Figure~\ref{fig:method}), which separates the nine ideal--actual use transitions into distinct
reward channels and uses
bidirectional counterfactual evidence to redistribute each channel's advantage
across response tokens while preserving its response-level mean.

\subsection{Ordered reward channels}

For each query--memory pair \((x,M)\), we sample a rollout group
\(\{y_i\}_{i=1}^{G}\), with
\(y_i\sim\pi_\theta(\cdot\mid x,M)\). Each atom's ideal use
level \(\levelideal_{n,k}\) is fixed across the group, whereas its actual use
level \(\levelhat_{n,k}(y_i)\) is assessed separately for each response \(y_i\).
For each \(y_i\), we categorize the atoms according to
\((\levelideal_{n,k},\levelhat_{n,k}(y_i))\) into nine ideal--actual reward
channels indexed by \(h\). To simplify notation, we denote
\ignorelevel{}, \boundlevel{}, and \controllevel{} by \(A\), \(B\), and
\(C\), respectively, and accordingly define a symbolic label for each of the
nine channels, as summarized in Table~\ref{tab:channels}. Let
\(\ell^{(h)}\) and \(\hat{\ell}^{(h)}\) denote the fixed
ideal and actual levels defining channel \(h\), and let
\(\mathcal I_i^{(h)}\) denote its atom-index set for response \(y_i\):
\[
\mathcal I_i^{(h)}
=\{(n,k):\levelideal_{n,k}=\ell^{(h)},\quad
\levelhat_{n,k}(y_i)=\hat{\ell}^{(h)}\}.
\]
To reward correct atom-level use and penalize incorrect use, we define each
channel reward as
\[
\begin{aligned}
R_i^{(h)}
&=\frac{\left|\mathcal I_i^{(h)}\right|}{Z^{(h)}}
\begin{cases}
+1,&\ell^{(h)}=\hat{\ell}^{(h)},\\
-\left|\operatorname{rank}(\hat{\ell}^{(h)})-
\operatorname{rank}(\ell^{(h)})\right|,&\ell^{(h)}\ne\hat{\ell}^{(h)},
\end{cases}\\
Z^{(h)}
&=\max\!\left(1,
\sum_{n=1}^{N}\sum_{k=1}^{K_n}
\mathbf 1[\levelideal_{n,k}=\ell^{(h)}]\right).
\end{aligned}
\]

\subsection{Bidirectional counterfactual localization}
\label{sec:localization}

Given the channel rewards, GRPO aggregates them before group normalization,
whereas GDPO preserves channel-specific signals by normalizing each channel
separately before aggregation \citep{liu2026gdpo}. Both still assign the
resulting response-level advantage uniformly to all tokens and therefore cannot
localize each channel's credit to the relevant tokens (see
Appendix~\ref{app:group-normalization} for details on both methods). In
\method{}, we use fixed-response counterfactual log-likelihood
differences to localize the influence of each channel's memory atoms and guide
advantage redistribution.

Specifically, for response \(y_i\), we separately localize each channel \(h\)
whose atom set satisfies \(\mathcal I_i^{(h)}\neq\varnothing\), except
\((\ignorelevel{},\ignorelevel{})\), which is excluded because correctly
ignored atoms leave no atom-specific response footprint. To localize channel
\(h\), we remove the atoms indexed by \(\mathcal I_i^{(h)}\), leaving all other
atoms and their order unchanged to obtain \(\widetilde{M}_i^{(h)}\). We then
perform teacher-forcing inference on the same generated response under
\(\widetilde{M}_i^{(h)}\) using the same rollout policy and define the
log-likelihood difference for its \(t\)-th token \(y_{i,t}\), conditioned on
the preceding tokens \(y_{i,<t}\), as
\[
d_{i,t}^{(h)}
=
\log p_\theta(y_{i,t}\mid x,M,y_{i,<t})
-
\log p_\theta
(y_{i,t}\mid x,\widetilde{M}_i^{(h)},y_{i,<t}).
\]
Since the response is fixed, \(d_{i,t}^{(h)}>0\) identifies tokens supported by
the removed atoms, whereas \(d_{i,t}^{(h)}<0\) identifies tokens suppressed by
them. We construct each channel's localization signal according to the
following rule: when
\(\operatorname{rank}(\hat{\ell}^{(h)})\ge
\operatorname{rank}(\ell^{(h)})\), except for
\((\ignorelevel{},\ignorelevel{})\), memory influence is realized or excessive,
so tokens with \(d_{i,t}^{(h)}>0\) serve as candidate locations. When
\(\operatorname{rank}(\hat{\ell}^{(h)})<
\operatorname{rank}(\ell^{(h)})\), actual use is insufficient. In this case,
tokens with \(d_{i,t}^{(h)}>0\) may reflect the portion of memory influence
already realized, so assigning the under-use penalty to them would suppress
correct behavior. Instead, we focus on tokens with \(d_{i,t}^{(h)}<0\), whose
likelihood the memory decreases but the model still generates, capturing cases
in which content remains in the response despite a suppressive memory. Correct
\((\ignorelevel{},\ignorelevel{})\) has zero localization signal because it
leaves no response footprint; its positive reward still contributes to
sequence-level credit. Accordingly, we define
\[
s^{(h)}=
\begin{cases}
0,&(\ell^{(h)},\hat{\ell}^{(h)})
    =(\ignorelevel{},\ignorelevel{}),\\
-1,&\operatorname{rank}(\hat{\ell}^{(h)})
    <\operatorname{rank}(\ell^{(h)}),\\
+1,&\operatorname{rank}(\hat{\ell}^{(h)})
    \ge\operatorname{rank}(\ell^{(h)})\ \land
    (\ell^{(h)},\hat{\ell}^{(h)})
    \ne(\ignorelevel{},\ignorelevel{}),
\end{cases}
\]
and \(z_{i,t}^{(h)}=s^{(h)}d_{i,t}^{(h)}\), so that positive
\(z_{i,t}^{(h)}\) marks the \(t\)-th token as a candidate location for the
credit of channel \(h\).
To make localization robust to extreme likelihood spikes and weak fluctuations,
we first clip \(z_{i,t}^{(h)}\) as
\(\bar z_{i,t}^{(h)}=\operatorname{clip}(z_{i,t}^{(h)},-d_{\max},d_{\max})\).
We then filter the localization signals at two levels: the
signal-strength gate
\(G_i^{(h)}=\mathbf 1[\max_t z_{i,t}^{(h)}>
\delta_{\mathrm{tok}}^{\mathrm{abs}}]\) suppresses localization when no
token-level signal exceeds the absolute threshold, while the channel-specific
threshold \(\delta^{(h)}\) filters out weak token-level fluctuations
in the clipped scores, yielding
\[
q_{i,t}^{(h)}=G_i^{(h)}
\operatorname{ReLU}(\bar z_{i,t}^{(h)}-\delta^{(h)}).
\]
We maintain each channel-specific threshold \(\delta^{(h)}\) using an
exponential moving average (EMA) (see Appendix~\ref{app:localization} for
details). The resulting \(q_{i,t}^{(h)}\) captures a token-level
manifestation of the atoms in channel \(h\) in the response and serves
as a proxy signal for fine-grained advantage redistribution.

\subsection{Channel-wise mean-preserving advantage redistribution}

Having obtained token-level localization signals for each localizable channel,
we normalize them within each response \(y_i\) of length \(T_i\):
\[
w_{i,t}^{(h)}=\frac{q_{i,t}^{(h)}}{Q_i^{(h)}},
\qquad
\text{where}\qquad
Q_i^{(h)}=\sum_{t=1}^{T_i} q_{i,t}^{(h)}.
\]
The normalized weight \(w_{i,t}^{(h)}\) determines the relative allocation of
the credit of channel \(h\) across response tokens. To redistribute the channel
advantage in these proportions while preserving the average advantage across
the response, we assign a multiplier to each token. Requiring the multipliers
to be proportional to \(w_{i,t}^{(h)}\) and average to one across tokens yields
\(m_{i,t}^{(h)}=T_iw_{i,t}^{(h)}\)
(see Appendix~\ref{app:redistribution} for the derivation and implementation
details).
For each channel
\(h\), the group-normalized advantage of response \(i\) is
\[
A_i^{(h)}=
\frac{R_i^{(h)}-\mu^{(h)}}{\sigma^{(h)}+\varepsilon},
\]
where \(\mu^{(h)}\) and \(\sigma^{(h)}\) are the mean and standard deviation of
the channel rewards across the rollout group. We then redistribute
\(A_i^{(h)}\) using the multiplier:
\[
\widetilde A_{i,t}^{(h)}
=A_i^{(h)}\left[(1-\eta_i^{(h)})+\eta_i^{(h)}m_{i,t}^{(h)}\right].
\]
Here \(\eta_i^{(h)}=\chi_i^{(h)}\eta^{(h)}\), where
\(\eta^{(h)}\in[0,1]\) is the channel-specific localization coefficient;
setting \(\eta^{(h)}\) to zero reduces the redistribution to uniform assignment
of \(A_i^{(h)}\) across response tokens. \(\chi_i^{(h)}\) is the redistribution
gate, defined as
\[
\chi_i^{(h)}=\mathbf 1\!\left[
R_i^{(h)}A_i^{(h)}>0\ \land\ Q_i^{(h)}>0\right],
\]
and permits token-level redistribution only when a nonzero localization
signal is available and the channel reward agrees in sign with the channel
advantage, preventing localized credit assignment from reversing the update
direction specified by the channel reward. Finally, we aggregate and normalize
the advantages across channels to maintain a stable learning scale and use the
resulting token advantages in the clipped GRPO objective to update the policy
(Appendix~\ref{app:redistribution}).

\section{Experiments}
\label{sec:experiments}

\subsection{Setup}

We compare \method{} with representative post-training methods, including SFT
\citep{ouyang2022training}, two OPSD variants (OPSD-PG and OPSD-GKD)
\citep{zhao2026selfdistilled}, GRPO \citep{shao2024deepseekmath}, and GDPO
\citep{liu2026gdpo}; we also report the original instruction-tuned models as
Base. From the 13,500 training examples, we hold out a stratified subset of
1,500 as a validation set for hyperparameter selection. For SFT, we fine-tune the model using
SFT data constructed from the remaining 12,000 examples
(see Appendix~\ref{app:sft-data} for construction details).
For OPSD, GRPO, GDPO, and \method{}, we initialize the model from the same
cold-start checkpoint trained on SFT data constructed from a 4,000-example
subset of the remaining 12,000 training examples and use the other 8,000
examples for subsequent training. To evaluate
\method{} across model families and scales, we conduct experiments on Qwen3-8B
\citep{yang2025qwen3}, Ministral-3-8B-Instruct \citep{liu2026ministral3}, and
Qwen3.5-35B-A3B \citep{qwenteam2026qwen35}. We evaluate
all models on the held-out 1,500-example test set following Section~\ref{sec:diagnosis},
reporting the mean and standard deviation over three seeds. Appendix~\ref{app:training-details} details the implementation of
all methods.

\subsection{Main results}

\begin{table*}[t]
\caption{Main results across model families and scales
(mean \(\pm\) standard deviation over three seeds). Red and blue denote the best
and second-best result for each model.}
\label{tab:main-results}
\begin{center}
\setlength{\tabcolsep}{2.8pt}
\begin{tabular}{lcccccc}
\toprule
\multirow{2}{*}{Method} & \multicolumn{2}{c}{Qwen3-8B} &
\multicolumn{2}{c}{Ministral-3-8B-Instruct} &
\multicolumn{2}{c}{Qwen3.5-35B-A3B}\\
\cmidrule(lr){2-3}\cmidrule(lr){4-5}\cmidrule(lr){6-7}
& SCS \(\uparrow\) & Exact \(\uparrow\) & SCS \(\uparrow\) &
Exact \(\uparrow\) & SCS \(\uparrow\) & Exact \(\uparrow\)\\
\midrule
Base & \(31.17{\pm}0.21\) & \(15.29{\pm}0.23\) &
\(31.88{\pm}0.58\) & \(16.84{\pm}0.48\) &
\(26.54{\pm}0.28\) & \(12.24{\pm}0.25\)\\
Cold-start & \(51.98{\pm}1.24\) & \(33.87{\pm}2.19\) &
\(66.87{\pm}1.05\) & \(50.62{\pm}1.56\) &
\(71.52{\pm}0.96\) & \(56.56{\pm}1.20\)\\
SFT & \(65.72{\pm}0.64\) & \(49.49{\pm}1.24\) &
\(70.94{\pm}0.28\) & \(55.89{\pm}0.52\) &
\(75.68{\pm}0.46\) & \(61.87{\pm}0.71\)\\
OPSD-PG & \(54.36{\pm}0.60\) & \(35.80{\pm}0.60\) &
\(70.48{\pm}0.42\) & \(54.71{\pm}0.30\) &
\(71.29{\pm}0.25\) & \(55.64{\pm}0.87\)\\
OPSD-GKD & \(57.39{\pm}0.41\) & \(39.49{\pm}0.38\) &
\(71.34{\pm}0.18\) & \(55.91{\pm}0.31\) &
\(71.67{\pm}0.85\) & \(56.89{\pm}1.55\)\\
GRPO & \(67.61{\pm}0.71\) & \(52.18{\pm}0.82\) &
\(\textcolor{blue}{77.21{\pm}0.46}\) & \(\textcolor{blue}{65.44{\pm}0.80}\) &
\(78.25{\pm}0.87\) & \(66.60{\pm}1.28\)\\
GDPO & \(\textcolor{blue}{72.25{\pm}0.30}\) &
\(\textcolor{blue}{57.91{\pm}0.40}\) & \(76.98{\pm}0.35\) &
\(63.84{\pm}0.43\) & \(\textcolor{blue}{79.39{\pm}1.02}\) &
\(\textcolor{blue}{67.51{\pm}1.44}\)\\
\method{} & \(\textcolor{red}{79.54{\pm}0.77}\) &
\(\textcolor{red}{67.89{\pm}1.07}\) & \(\textcolor{red}{79.00{\pm}0.19}\) &
\(\textcolor{red}{66.31{\pm}0.60}\) & \(\textcolor{red}{81.12{\pm}0.78}\) &
\(\textcolor{red}{70.16{\pm}1.23}\)\\
\bottomrule
\end{tabular}
\end{center}
\end{table*}

As shown in Table~\ref{tab:main-results}, \method{} achieves the highest SCS
and Exact across all three models, with improvements over the strongest
baseline ranging from 1.73 to 7.29 on SCS and from 0.87 to 9.98 on Exact. The
complete directional results in Table~\ref{tab:full-main-results} show that,
relative to Cold-start, all other post-training methods reduce errors in one
direction while increasing errors in the other on at least one model. \method{}
is the only method that reduces both over-use and under-use across all three
models, demonstrating the effectiveness of using bidirectional counterfactual
evidence to guide advantage redistribution across model families and scales.
Moreover, without additional training, \method{}
achieves the strongest overall performance on RPEval, showing that the gains
transfer to an external memory-use benchmark
(Appendix~\ref{app:rpeval-results}).

\subsection{Ablations and robustness}

\begin{figure*}[t]
\centering
\includegraphics[width=\textwidth]{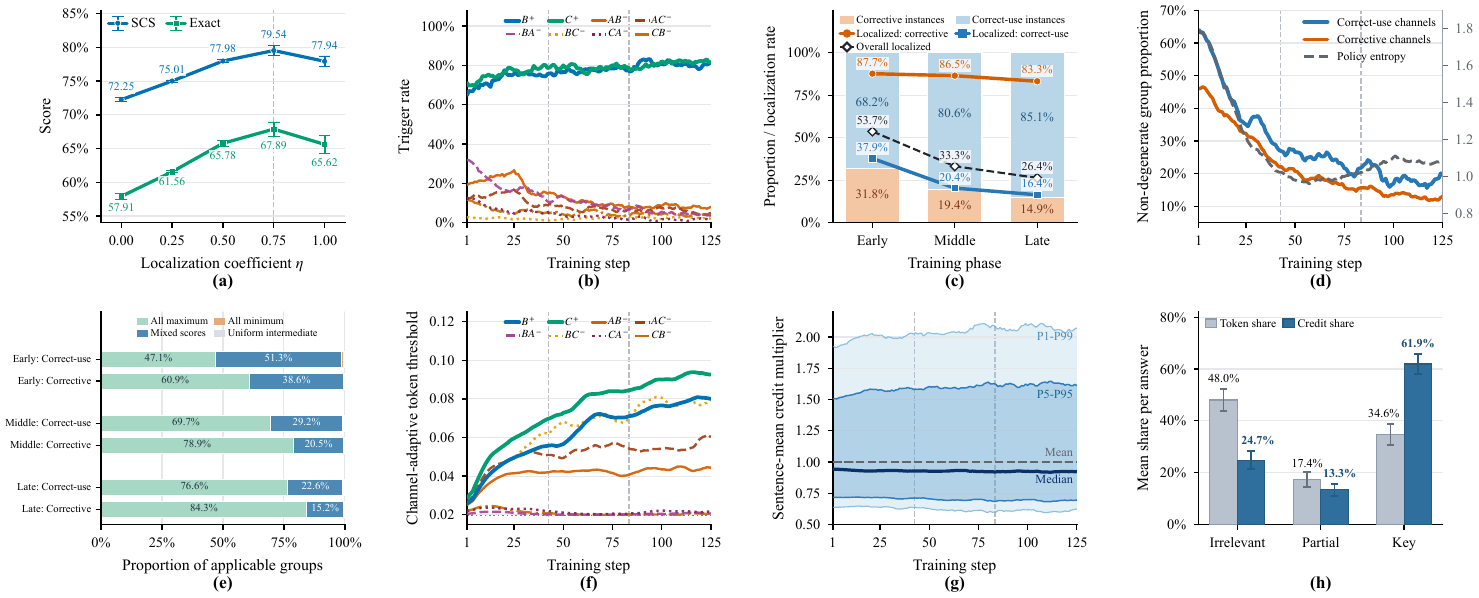}
\caption{Sensitivity and mechanism analyses on Qwen3-8B. (a) Sensitivity to
the localization coefficient \(\eta\); (b--g) training dynamics of \method{};
and (h) token and
credit shares across human-annotated sentence roles for all eight localizable
channels. Curves in (b), (d), (f),
and (g) use a seven-step centered mean; error bars in (a) show standard
deviations over three seeds and those in (h) show 95\% bootstrap
confidence intervals.}
\label{fig:mechanism-analysis}
\end{figure*}

Having established the overall gains of \method{}, we next examine its
localization design and the robustness of the resulting comparison across
Judges. To this end, we conduct three
complementary experiments on Qwen3-8B. First, to isolate the effects of
localization granularity and direction-aware counterfactual evidence, we ablate
both design choices. Sentence-level localization serves as a natural
coarse-grained alternative: it segments the response into sentence units and
applies one multiplier to all tokens within each unit, whereas token-level
localization distinguishes individual tokens. At each granularity, we compare
ordered bidirectional localization with an absolute-magnitude variant that
retains only the magnitude of the counterfactual likelihood change
(implementation details in Appendix~\ref{app:localization-ablations}). As shown in
Table~\ref{tab:localization-ablation}, ordered bidirectional localization
improves SCS and Exact by 1.59 and 1.80 points at the sentence level and by
4.84 and 6.76 points at the token level, with token-level ordered localization
achieving the highest SCS and Exact scores. These results support our design in \method{},
which combines token-level
localization with ordered bidirectional counterfactual evidence.
Second, to assess sensitivity to localization strength, we evaluate
\(\eta^{(h)}\in\{0,0.25,0.5,0.75,1.0\}\) as the shared localization coefficient
across the eight localizable channels, while holding all other settings fixed.
Figure~\ref{fig:mechanism-analysis}(a)
shows that SCS and Exact both peak at \(\eta^{(h)}=0.75\), reaching 79.54 and
67.89, respectively. These results suggest that retaining a sequence-level component alongside
token-level localization is beneficial. Further details are
provided in Appendix~\ref{app:eta-sensitivity}.

Finally, alongside the human evaluation of DeepSeek-V4-Pro in
Appendix~\ref{app:judge-audit}, we re-evaluate the Qwen3-8B results in
Table~\ref{tab:main-results} using Qwen3.8-Max
\citep{qwenteam2026qwen38max} as an alternative Judge.
The results in Table~\ref{tab:alternative-judge-results} show that \method{}
retains the highest SCS and Exact and the lowest \smus{} and \aur{} among
the eight methods.
Method rankings also remain highly
consistent with those obtained using DeepSeek-V4-Pro, with Spearman
correlations of 0.976 for SCS and 1.000 for Exact
(Table~\ref{tab:judge-stability}). These results show that \method{}'s
advantage in overall memory use extends to evaluation by an alternative
Judge. Appendix~\ref{app:judge-stability}
provides the full results and comparison procedure.

\begin{table}[ht]
\caption{Localization ablations on Qwen3-8B (mean \(\pm\) standard deviation over three seeds). 
Red and blue denote the best and second-best
result in each column.}
\label{tab:localization-ablation}
\begin{center}
\setlength{\tabcolsep}{1.5pt}
\begin{tabular}{lccccccc}
\toprule
Granularity & Signal & SCS \(\uparrow\) & Exact \(\uparrow\) & \smos{} \(\downarrow\) &
\smus{} \(\downarrow\) & \aor{} \(\downarrow\) & \aur{} \(\downarrow\)\\
\midrule
Sentence & Absolute & \(76.59{\pm}0.11\) &
\(63.87{\pm}0.27\) & \(\textcolor{blue}{13.70{\pm}0.45}\) &
\(11.65{\pm}0.27\) & \(\textcolor{blue}{21.53{\pm}0.75}\) &
\(19.80{\pm}0.53\)\\
Sentence & Ordered & \(\textcolor{blue}{78.18{\pm}0.04}\) &
\(\textcolor{blue}{65.67{\pm}0.53}\) &
\(\textcolor{red}{12.98{\pm}0.19}\) &
\(\textcolor{blue}{10.52{\pm}0.18}\) &
\(\textcolor{red}{20.60{\pm}0.55}\) &
\(\textcolor{blue}{18.20{\pm}0.44}\)\\
Token & Absolute & \(74.70{\pm}0.78\) &
\(61.13{\pm}0.98\) & \(14.40{\pm}0.39\) & \(12.96{\pm}0.48\) &
\(22.20{\pm}0.55\) & \(22.11{\pm}0.68\)\\
Token & Ordered &
\(\textcolor{red}{79.54{\pm}0.77}\) &
\(\textcolor{red}{67.89{\pm}1.07}\) & \(14.98{\pm}0.60\) &
\(\textcolor{red}{6.81{\pm}0.28}\) & \(23.49{\pm}0.74\) &
\(\textcolor{red}{12.29{\pm}0.57}\)\\
\bottomrule
\end{tabular}
\end{center}
\end{table}

\subsection{Mechanism analysis}
We then assess whether counterfactual localization directs credit to
memory-influenced tokens and analyze the training dynamics of \method{}. For
localization analysis, we sample 20 response--channel pairs from each of the
eight localizable channels. For each sampled pair,
we first aggregate token-level credit within each sentence, and a human
annotator then labels each sentence as \textsc{Irrelevant}, \textsc{Partial},
or \textsc{Key}. Using these annotations, we compute, for each pair, the
fraction of response tokens contained in sentences of each label and the
fraction of total credit assigned to those sentences, and record
a Top-3 localization hit if at least one of the three highest-credit sentences
is labeled \textsc{Partial} or \textsc{Key}. Across all eight channels,
Figure~\ref{fig:mechanism-analysis}(h) shows that
\textsc{Irrelevant}, \textsc{Partial}, and \textsc{Key} sentences account for
48.0\%, 17.4\%, and 34.6\% of response tokens but receive 24.7\%, 13.3\%, and
61.9\% of the total credit, respectively. The Top-3 hit rate is 99.0\% for
the five support-targeting channels (\(AB^-\), \(AC^-\), \(B^+\), \(BC^-\),
and \(C^+\)) and 96.7\% for the three under-use channels
(\(BA^-\), \(CA^-\), and \(CB^-\)). These results show that counterfactual
localization indeed assigns more credit to memory-influenced content in the
response. Annotation and aggregation details,
together with representative cases from both channel groups, are provided in
Appendix~\ref{app:localization-audit}.

To analyze training dynamics, we divide the eight localizable channels into a
correct-use group (\(B^+\) and \(C^+\)) and a corrective group (the other six).
Figure~\ref{fig:mechanism-analysis}(b) shows that the \(B^+\) and \(C^+\)
trigger rates rise, whereas five of the six corrective channels decline, with
the low-frequency \(BC^-\) channel remaining stable. This trend indicates that
the policy progressively shifts from memory-use errors toward correct use.
Figure~\ref{fig:mechanism-analysis}(c) shows that the token-level
localization rate falls from 37.9\% to 16.4\% for correct-use channels but only
from 87.7\% to 83.3\% for corrective channels.
Figure~\ref{fig:mechanism-analysis}(d) explains the declining localization
rates: the non-degenerate-group rates of both channel groups decrease, showing
that the rollouts for a query increasingly receive identical channel
rewards. As a result, the redistribution gate more frequently falls back to
sequence-level credit. Although policy entropy initially falls in
parallel, it recovers late while the non-degenerate-group rates remain low,
showing that the loss of within-group reward variation reflects convergence in
memory-use outcomes rather than entropy collapse.
Figure~\ref{fig:mechanism-analysis}(e) further shows that group degeneration
mainly reflects uniformly correct memory use in correct-use channels and the
uniform absence of the corresponding error in corrective channels. Together,
Figures~\ref{fig:mechanism-analysis}(b--e) show that training yields more
appropriate and consistent memory use, while the mechanism correctly falls
back when comparative signals vanish.
Finally,
Figures~\ref{fig:mechanism-analysis}(f--g) show that channel thresholds separate
into distinct empirical scales, supporting channel-adaptive filtering, while
the token-weighted mean sentence multiplier remains 1 as its distribution
widens, confirming differentiated local credit without response-level scale
drift. Full analysis details and supporting statistics for
Figure~\ref{fig:mechanism-analysis}(b--g) are provided in
Appendix~\ref{app:training-dynamics}.

\section{Conclusion and Limitations}

Agent memory is effective only if the LLM gives each supplied
proposition the appropriate level of influence. Our results show that this
capability cannot be taken for granted: frontier models frequently mismatch
target and actual use levels, while common post-training algorithms can improve
one error direction at the expense of the other. \dataset{} makes this
overlooked problem measurable, while \method{} addresses it through
bidirectional counterfactual credit localization and advantage redistribution,
yielding stronger overall memory-use performance and a better balance between
over-use and under-use.

Counterfactual localization provides a proxy for locating channel-level credit
within a response, but its token-level attribution may not always be accurate.
However, the algorithm's robust design mitigates this limitation through
sequence-level fallback and channel-specific control of localization strength
via \(\eta^{(h)}\); experiments across model families and scales, together
with human evaluation, demonstrate the practical effectiveness of counterfactual
localization. Developing more principled and accurate ways to use this signal
remains an important direction for future work.

\section*{Acknowledgments}

This work was supported by Qwen Business Unit through Alibaba Research Intern Program.

\bibliography{references}
\bibliographystyle{abbrvnat}

\clearpage
\appendix

\section{Related Work}
\label{app:related-work}
\paragraph{Agent memory and evaluation.}
Agent memory research focuses on the formation, updating, and retrieval of
factual, experiential, and working memory
\citep{hu2026agentmemorysurvey,luo2026storage}.
LongMemEval and LoCoMo evaluate long-term conversational memory and reasoning
\citep{wu2024longmemeval,maharana2024locomo}.
MemoryAgentBench assesses retrieval, test-time learning, long-range
understanding, and selective forgetting through incremental interactions
\citep{hu2026memoryagentbench}; Mem2ActBench evaluates memory-driven tool
selection and parameter grounding \citep{shen2026mem2act}.
Memory-R1 uses reinforcement learning to optimize memory management and
memory-based answer generation \citep{yan2026memoryr1}, while ReMe distills,
reuses, and refines procedural experience \citep{cao2026reme}.

\paragraph{Selective and personalized memory use.}
OP-Bench and MemSyco-Bench expose over-personalization and memory-induced
sycophancy \citep{hu2026opbench,xiang2026memsyco}.
RPEval assigns user preferences \textsc{Ignore}, \textsc{Support}, or
\textsc{Dominate} strategies and evaluates personalization errors arising from
mismatches between intended and actual use \citep{feng2026rpeval}.
BenchPreS evaluates whether stored preferences are appropriately applied or
suppressed in third-party communication contexts \citep{yoon2026benchpres}.
StratMem-Bench distinguishes required, supportive, and irrelevant memories in
virtual-character dialogue and assesses item-level use \citep{wu2026stratmem}.
\dataset{} evaluates whether models use memory appropriately across health,
general assistance, and coding, with atomic propositions embedded in natural composite memory
blocks. Atom-specific rubrics assess each proposition's actual influence on
the response against its target level, yielding measures of over-use,
under-use, and overall memory-use performance.
Table~\ref{tab:benchmark-comparison} summarizes the evaluation scope and
granularity of these benchmarks.

\begin{table}[htbp]
\caption{Comparison of memory-use benchmarks by evaluation scope and
granularity. $\checkmark$ indicates an explicit evaluation task or criterion;
$\times$ indicates that the aspect is not explicitly evaluated. Conflict
handling covers conflicts with task requirements, facts, or other memories.
Item-level evaluation assesses individual memories or preferences.}
\label{tab:benchmark-comparison}
\begin{center}
\normalsize
\setlength{\tabcolsep}{2pt}
\renewcommand{\arraystretch}{1.2}
\begin{tabular}{@{}>{\raggedright\arraybackslash}m{0.22\linewidth}
>{\raggedright\arraybackslash}m{0.18\linewidth}
>{\centering\arraybackslash}m{0.12\linewidth}
*{2}{>{\centering\arraybackslash}m{0.10\linewidth}}
>{\raggedright\arraybackslash}m{0.22\linewidth}@{}}
\toprule
Benchmark & Setting & Applica\-bility & Conflict handling & Per-item use
& Memory-use evaluation unit \\
\midrule
OP-Bench\newline \citep{hu2026opbench} & Personalized dialogue
& $\checkmark$ & $\checkmark$ & $\times$
& Response; response set \\
MemSyco-Bench\newline \citep{xiang2026memsyco} & Personalized QA and decisions
& $\checkmark$ & $\checkmark$ & $\times$ & Response \\
RPEval\newline \citep{feng2026rpeval} & Everyday assistance
& $\checkmark$ & $\checkmark$ & $\checkmark$
& Individual preference \\
BenchPreS\newline \citep{yoon2026benchpres} & Third-party communication
& $\checkmark$ & $\checkmark$ & $\checkmark$
& Preference attribute \\
StratMem-Bench\newline \citep{wu2026stratmem} & Virtual-character dialogue
& $\checkmark$ & $\times$ & $\checkmark$
& Individual memory item \\
\midrule
\dataset{} (ours) & Health, general assistance, coding
& $\checkmark$ & $\checkmark$ & $\checkmark$
& Atomic proposition within a memory block \\
\bottomrule
\end{tabular}
\end{center}
\end{table}
\FloatBarrier

\section{\dataset{} Construction and Data Statistics}
\label{app:data}

\subsection{Construction and quality control}
\label{app:construction-process}

\begin{figure}[t]
\centering
\includegraphics[width=\textwidth]{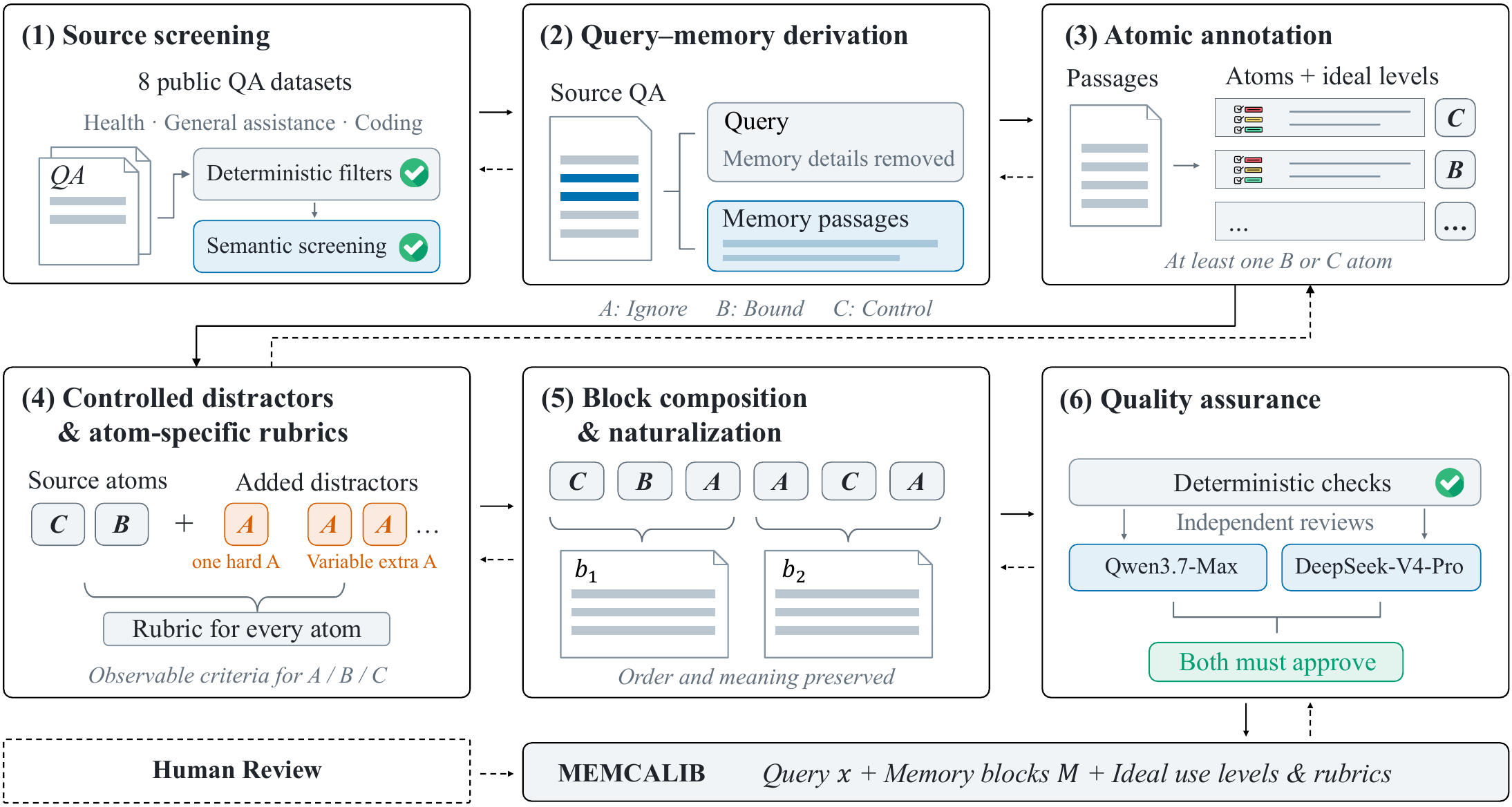}
\caption{Overview of the \dataset{} construction pipeline and its refinement
through human feedback. During pipeline development, each iteration constructs
a fresh batch of 50 pilot samples balanced across the three
domains. Human review identifies sample-level issues and guides revisions
to the relevant stages and prompts. This process continues until the review
pass rate reaches at least 90\%, after which the pipeline is frozen for
full-scale construction. Solid arrows indicate the sequence of data construction
steps, while dashed arrows indicate human feedback for pipeline refinement.}
\label{fig:data}
\end{figure}

We construct \dataset{} from eight public question-answering datasets spanning
health dialogue, general assistance, and coding. Table~\ref{tab:source-datasets}
lists the source datasets and summarizes the tasks they cover in each domain.
We derive each sample using a six-stage pipeline that is iteratively refined
based on human review of fresh pilot batches and then frozen for full-scale
construction. Figure~\ref{fig:data} illustrates both the six-stage pipeline
and its refinement through human feedback. During data construction, we
use Qwen3.7-Max \citep{qwenteam2026qwen37max} with stage-specific prompts for
generation and annotation.

\begin{table}[h]
\caption{Public datasets used to construct \dataset{}.}
\label{tab:source-datasets}
\begin{center}
\setlength{\tabcolsep}{3pt}
\begin{tabular}{>{\raggedright\arraybackslash}p{0.19\linewidth}
>{\raggedright\arraybackslash}p{0.42\linewidth}
>{\raggedright\arraybackslash}p{0.30\linewidth}}
\toprule
Domain & Source datasets & Coverage\\
\midrule
Health dialogue & MedDialog \citep{he2020meddialog};
ChatDoctor--HealthCareMagic \citep{li2023chatdoctor} & Symptoms, medication,
chronic conditions, and care seeking\\
General assistance & UltraChat 200K \citep{ding2023ultrachat};
OpenAssistant OASST1 and OASST2 \citep{kopf2023openassistant} & Explanation,
planning, writing, and everyday assistance\\
Coding & Stack Exchange Preferences \citep{lambert2023stackexchange};
APPS \citep{hendrycks2021apps}; Magicoder OSS-Instruct \citep{wei2023magicoder} & Algorithms,
APIs and libraries, systems, and debugging\\
\bottomrule
\end{tabular}
\end{center}
\end{table}

\paragraph{1. Source screening.}
We first assess source samples for query--memory derivation through
deterministic filtering followed by LLM-based semantic screening.
Deterministic filtering checks that sample identifiers, source information,
questions, and reference answers are present, enforces English-language
and length requirements, and removes noisy text and exact or near-duplicate
samples. Semantic screening checks whether the source question and reference
answer form a coherent, informative QA pair, and whether the question
or context contains explicit information suitable for memory extraction.

\paragraph{2. Query and memory derivation.}
For each admitted source sample, we prompt the LLM to extract facts,
preferences, constraints, and past events explicitly stated in the question
or context, and group related details into memory passages for subsequent
atomic annotation. The LLM rewrites the question to preserve the original
request while removing the information assigned to these passages.
The reference answer aids task understanding during this process.
During generation, we instruct the LLM to produce a query that neither
states nor implies any extracted memory detail and remains coherent
and answerable with the extracted memories. We verify both conditions
during the subsequent independent LLM-based review described in Stage 6.

\paragraph{3. Atomic annotation.}
We next use the LLM to decompose the extracted memory passages into
independently judgeable atomic propositions and assign each atom an ideal
use level: no answer-specific footprint (\ignorelevel{}), bounded local
support (\boundlevel{}), or control over a material conclusion, constraint,
or recommendation (\controllevel{}). To ensure that each sample tests the
use of task-relevant memory, we retain only samples containing at least
one \boundlevel{} or \controllevel{} atom.

\paragraph{4. Controlled distractor construction.}
Because source questions and their context primarily contain information
relevant to the original request, the LLM assigns most extracted atoms a
\boundlevel{} or \controllevel{} level in Stage 3. To assess models' ability
to ignore distracting memories, we prompt the LLM to generate one hard
\ignorelevel{} atom and a variable number of additional \ignorelevel{} atoms
for each retained item. The LLM constructs the hard atom to appear relevant
through topical proximity or same-user plausibility while keeping its content
outside the current answer scope. It generates the additional atoms to vary
memory-block length and composition and simulate long-tail retrieval noise.
After adding these distractors, we prompt the LLM to generate an atom-specific
rubric for every source-derived and added atom. Each rubric specifies
observable response-text criteria for distinguishing the three actual use
levels.

\paragraph{5. Block composition and naturalization.}
After adding the distractors and generating a rubric for each atom, we
assemble the atoms into memory blocks using predefined counts for the blocks
in each sample and the atoms within each block. These counts vary across
samples, with atom counts also varying across blocks within each sample, to
cover different memory-context sizes and block complexities.
We preserve atom order and allow
atoms with different ideal use levels to coexist within a block, to reflect
how a retrieved memory can mix useful context with irrelevant details.
We then prompt the LLM to rewrite each block as a coherent natural-language
paragraph, preserving every proposition's meaning without omitting atoms or
introducing additional information.

\paragraph{6. Quality assurance.}
After naturalizing the memory blocks, we validate each completed sample.
We first run deterministic checks to verify that each sample contains the
query, memory blocks, and atomic annotations, that each atom has a valid ideal
use level and a complete rubric, and that every atom belongs to exactly one
block in its recorded order. We then prompt Qwen3.7-Max and DeepSeek-V4-Pro
\citep{deepseekai2026deepseekv4} to review each sample independently. Each LLM
checks whether source-derived atoms faithfully capture the original context
and express distinct propositions; whether the query excludes extracted
memory details and remains coherent and answerable with the memories; whether
rewritten blocks preserve every proposition without adding information; and
whether ideal levels reflect each atom's role in the task and rubrics provide
observable response-text criteria for distinguishing actual use levels.
We accept a sample through semantic review only when both LLMs approve it.

\paragraph{Iterative refinement through human feedback.}
Before full-scale construction, we iteratively refine the six-stage pipeline
through human review. At each iteration, we construct a fresh batch of 50 pilot
samples balanced across health, general assistance, and
coding. We manually
inspect all 50 pilot samples, including their query--memory splits, ideal use levels, and
rubric criteria. A sample passes only if no issues are identified across all
reviewed aspects. For samples that fail, we analyze the identified issues,
trace them to the relevant construction stages, and revise the corresponding
procedures and LLM prompts. We repeat this process with a newly constructed batch
at each iteration until the sample-level human-review pass rate reaches at
least 90\%. We then freeze the pipeline and use it for full-scale benchmark
construction.

The \hyperref[ex:benchmark-example]{example below} illustrates a constructed
health-domain sample, including its query, memory blocks, ideal use levels,
and rubric criteria for assessing each atom's influence on the response.
For space, we show only selected rubric fields for one atom at each ideal
use level.
\begin{benchmarkexamplebox}[nameref={Benchmark example},phantomlabel={ex:benchmark-example}]
\normalsize
\textbf{Current query \(x\)}

What medical evaluations are typically recommended for a couple who has been having frequent unprotected intercourse without conceiving?

\medskip
\textbf{Memory blocks \(M\)}

\smallskip
\(b_{1}\)\quad The user is 21 years old and has been attempting to conceive through daily unprotected intercourse for three years. The user's partner is 24 years old, and the couple resides in a suburban apartment complex.

\smallskip
\(b_{2}\)\quad The user reports irregular menstrual cycles and tracks menstrual cycles using a paper calendar.

\smallskip
\(b_{3}\)\quad The user is currently writing a master's thesis on the sociological impacts of infertility treatment access.

\smallskip
\(b_{4}\)\quad The user acquired an indigo coin display tray and stored it in the foyer sideboard drawer in May 2008, having lined the tray with acid-free felt before placing it in that drawer in May 2008.

\medskip
\textbf{Evaluation annotations (not provided to the model)}

\smallskip
{
\setlength{\tabcolsep}{4pt}
\renewcommand{\arraystretch}{1.12}
\begin{tabular}{@{}>{\raggedright\arraybackslash}p{0.09\linewidth}>{\raggedright\arraybackslash}p{\dimexpr0.76\linewidth-4\tabcolsep\relax}>{\raggedright\arraybackslash}p{0.15\linewidth}@{}}
\toprule
Atom & Atomic proposition & Ideal level\\
\midrule
\(m_{1,1}\) & The user is 21 years old. & \boundlevel{} \\
\addlinespace[4pt]
\(m_{1,2}\) & The user has been attempting to conceive through daily unprotected intercourse for three years. & \controllevel{} \\
\addlinespace[4pt]
\(m_{1,3}\) & The user's partner is 24 years old. & \ignorelevel{} \\
\addlinespace[4pt]
\(m_{1,4}\) & The couple resides in a suburban apartment complex. & \ignorelevel{} \\
\addlinespace[4pt]
\(m_{2,1}\) & The user reports irregular menstrual cycles. & \controllevel{} \\
\addlinespace[4pt]
\(m_{2,2}\) & The user tracks menstrual cycles using a paper calendar. & \ignorelevel{} \\
\addlinespace[4pt]
\(m_{3,1}\) & The user is currently writing a master's thesis on the sociological impacts of infertility treatment access. & \ignorelevel{} \\
\addlinespace[4pt]
\(m_{4,1}\) & The user acquired an indigo coin display tray and stored it in the foyer sideboard drawer in May 2008. & \ignorelevel{} \\
\addlinespace[4pt]
\(m_{4,2}\) & The indigo coin display tray was lined with acid-free felt before being placed in the foyer sideboard drawer in May 2008. & \ignorelevel{} \\
\bottomrule
\end{tabular}}

\medskip
\textbf{Rubric examples}

\medskip
\noindent\begin{minipage}{\linewidth}
\textbf{\(m_{3,1}\) --- Ideal level: \ignorelevel{}}
\par\smallskip
\textit{Expected behavior.} Answer provides a direct, clinically focused list of evaluations without referencing the user's academic work or sociological perspectives.
\par\smallskip
\textit{Over-use.} Inclusion of academic framing, sociological terminology, or references to the thesis in the medical advice.
\end{minipage}

\medskip
\noindent\begin{minipage}{\linewidth}
\textbf{\(m_{1,1}\) --- Ideal level: \boundlevel{}}
\par\smallskip
\textit{Expected behavior.} Answer references user's young age to contextualize evaluation recommendations or prognosis without altering core test list.
\par\smallskip
\textit{Under-use.} No acknowledgment of age despite its relevance to fertility counseling.
\par\smallskip
\textit{Over-use.} Attributing infertility solely to age or recommending different tests based only on age.
\end{minipage}

\medskip
\noindent\begin{minipage}{\linewidth}
\textbf{\(m_{1,2}\) --- Ideal level: \controllevel{}}
\par\smallskip
\textit{Expected behavior.} Answer explicitly recommends medical fertility evaluation for both partners based on duration of unprotected intercourse.
\par\smallskip
\textit{Under-use.} Failing to recommend evaluation despite three-year history.
\end{minipage}
\end{benchmarkexamplebox}

\subsection{Benchmark composition}
\label{app:benchmark-composition}

\dataset{} contains 15,000 examples across health, general assistance,
and coding. Table~\ref{tab:benchmark-composition} reports the total numbers of
examples, memory blocks, and atomic propositions in each domain, while
Table~\ref{tab:benchmark-structure} details the distribution of ideal use levels
and the structure of the memory blocks. A mixed-target block contains atoms
with at least two different ideal use levels. The percentages indicate the
share of atoms at each ideal level and the share of examples containing a
mixed-target block. The ranges give the minimum and maximum numbers of atoms
per block and per example. To support consistent response-text evaluation
across domains, we formulate coding queries as natural-language tasks covering
implementation planning, behavior prediction, code understanding, and
debugging diagnosis, allowing us to apply atom-specific rubrics directly to
the generated responses.

\begin{table}[h]
\caption{\dataset{} composition by domain.
Block and proposition counts are totals over examples in each domain.}
\label{tab:benchmark-composition}
\begin{center}
\begin{tabular}{lrrr}
\toprule
Domain & Examples & Memory blocks & Atomic propositions\\
\midrule
Health & 7,500 & 37,400 & 117,110\\
General assistance & 3,750 & 18,700 & 58,495\\
Coding & 3,750 & 18,700 & 58,615\\
\midrule
Total & 15,000 & 74,800 & 234,220\\
\bottomrule
\end{tabular}
\end{center}
\end{table}

\begin{table}[h]
\caption{Ideal use levels and memory structure in \dataset{}.}
\label{tab:benchmark-structure}
\begin{center}
\begin{tabular}{lr}
\toprule
Statistic & Value\\
\midrule
Atoms with ideal level \ignorelevel{} & 197,510 (84.3\%)\\
Atoms with ideal level \boundlevel{} & 18,943 (8.1\%)\\
Atoms with ideal level \controllevel{} & 17,767 (7.6\%)\\
Mixed-target blocks & 20,097\\
Examples containing a mixed-target block & 14,949 (99.66\%)\\
Atoms per block (min--max) & 1--20\\
Atoms per example (min--max) & 6--63\\
\bottomrule
\end{tabular}
\end{center}
\end{table}

\section{Human Evaluation of the Judge}
\label{app:judge-audit}
To assess how reliably DeepSeek-V4-Pro judges the actual use level of each
memory atom in a model response, we sample 300 response--atom pairs for human
evaluation. Of these, 150 are randomly sampled to measure agreement under the
natural evaluation distribution. The other 150 are stratified across the nine
combinations of ideal and Judge-assessed actual use levels to examine less
frequent combinations and use-level boundaries. For each pair, a human annotator applies the same atom-specific
rubric, without seeing the Judge's decision, to assess its actual use level as
\ignorelevel{}, \boundlevel{}, or \controllevel{}.
\begin{figure}[t]
\centering
\includegraphics[width=0.9\textwidth]{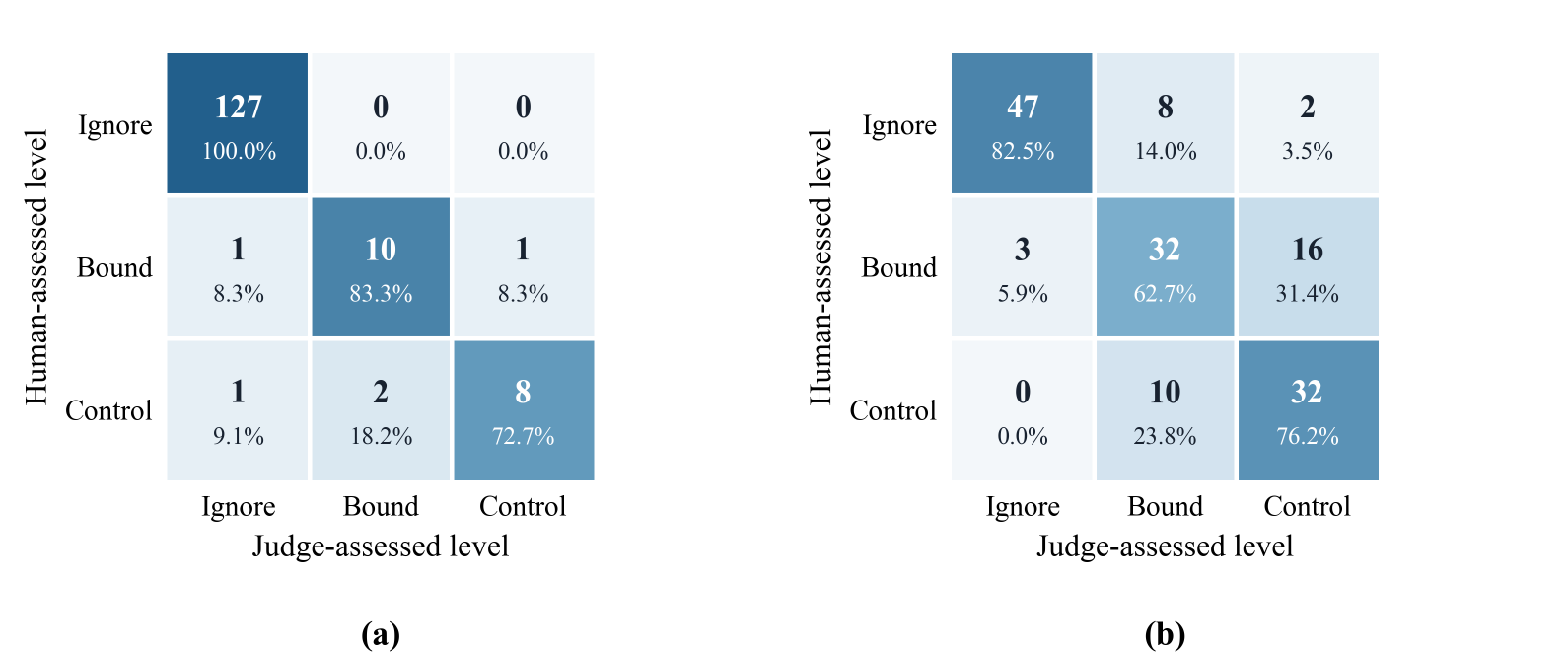}
\caption{Confusion matrices comparing human and DeepSeek-V4-Pro assessments of
actual use levels. Rows correspond to human assessments and columns to the
Judge's assessments; cells show counts and row-normalized percentages.
(a) Natural-distribution subset. (b) Stress-stratified subset.}
\label{fig:judge-human-audit}
\end{figure}
The results in Figure~\ref{fig:judge-human-audit}(a) show strong agreement under
the natural evaluation distribution: the Judge and human annotator assess the
same actual use level in 145 of 150 cases, yielding 96.7\% exact agreement
(95\% Wilson CI: 92.4\%--98.6\%), with Cohen's \(\kappa=0.872\). All 127
human-assessed \ignorelevel{} cases match the Judge's assessment. Among the five
disagreements, three lie at the \boundlevel{}/\controllevel{} boundary; the
Judge assesses a lower actual use level in four cases and a higher one in one.
The stress-stratified results in Figure~\ref{fig:judge-human-audit}(b) yield
111/150 exact agreement (74.0\%; 95\% Wilson CI: 66.4\%--80.4\%), with Cohen's
\(\kappa=0.610\). The largest concentration of disagreements again lies at the
\boundlevel{}/\controllevel{} boundary: 16 cases are assessed as \boundlevel{}
by the human annotator and \controllevel{} by the Judge, with 10 in the reverse
direction. Together, these results support the reliability of DeepSeek-V4-Pro
as the Judge under the natural evaluation distribution, while identifying the
distinction between bounded and controlling influence as the primary remaining
uncertainty.

\section{\method{} Details}
\label{app:method}

\subsection{Reward channels and localization targets}
\label{app:channels}

Table~\ref{tab:channels} summarizes the notation, per-atom rewards, and
localization targets for all nine ideal--actual use transitions.
The superscripts \(+\) and \(-\) in the channel labels indicate the sign of
the reward. The sign of \(d\) indicates whether the channel's memory atoms
increase or decrease the likelihood of a generated token relative to removing
those atoms.

\begin{table}[ht]
\caption{Nine ideal--actual reward channels, shorthand notation, per-atom
rewards, and localization targets.}
\label{tab:channels}
\begin{center}
\setlength{\tabcolsep}{4.5pt}
\begin{tabular}{cccc}
\toprule
Notation & Ideal (\(\ell^{(h)}\)) \(\rightarrow\)
actual (\(\hat{\ell}^{(h)}\)) & Per-atom reward &
Localization target\\
\midrule
\(A^+\) & \ignorelevel{}\(\rightarrow\)\ignorelevel{} &
\(+1\) & sequence-level\\
\(AB^-\) & \ignorelevel{}\(\rightarrow\)\boundlevel{} &
\(-1\) & supported tokens (\(d>0\)) or fallback\\
\(AC^-\) & \ignorelevel{}\(\rightarrow\)\controllevel{} &
\(-2\) & supported tokens (\(d>0\)) or fallback\\
\(BA^-\) & \boundlevel{}\(\rightarrow\)\ignorelevel{} &
\(-1\) & suppressed tokens (\(d<0\)) or fallback\\
\(B^+\) & \boundlevel{}\(\rightarrow\)\boundlevel{} &
\(+1\) & supported tokens (\(d>0\)) or fallback\\
\(BC^-\) & \boundlevel{}\(\rightarrow\)\controllevel{} &
\(-1\) & supported tokens (\(d>0\)) or fallback\\
\(CA^-\) & \controllevel{}\(\rightarrow\)\ignorelevel{} &
\(-2\) & suppressed tokens (\(d<0\)) or fallback\\
\(CB^-\) & \controllevel{}\(\rightarrow\)\boundlevel{} &
\(-1\) & suppressed tokens (\(d<0\)) or fallback\\
\(C^+\) & \controllevel{}\(\rightarrow\)\controllevel{} &
\(+1\) & supported tokens (\(d>0\)) or fallback\\
\bottomrule
\end{tabular}
\end{center}
\end{table}

\subsection{Group normalization and policy objective}
\label{app:group-normalization}

GRPO aggregates channel rewards before group normalization, whereas GDPO
normalizes each channel separately before aggregation.
For a rollout group of \(G\) responses to the same query--memory pair
\((x,M)\), GRPO computes the total reward \(R_i=\sum_h R_i^{(h)}\) for the
\(i\)-th response and normalizes it within the group \citep{shao2024deepseekmath}:
\[
A_i^{\mathrm{GRPO}}
=\frac{R_i-\mu_R}{\sigma_R+\varepsilon},
\]
where \(\mu_R\) and \(\sigma_R\) are the mean and standard deviation of
the total rewards across the \(G\) responses:
\[
\mu_R=\frac{1}{G}\sum_{i=1}^{G}R_i,
\qquad
\sigma_R=\sqrt{\frac{1}{G}\sum_{i=1}^{G}(R_i-\mu_R)^2}.
\]
Here \(\varepsilon>0\) is a small constant included in the denominators to
avoid division by zero. GRPO then assigns this response-level advantage
uniformly to all tokens in \(y_i\):
\[
A_{i,t}^{\mathrm{GRPO}}=A_i^{\mathrm{GRPO}},
\qquad t=1,\ldots,T_i,
\]
where \(T_i\) is the length of response \(y_i\).

GDPO first normalizes each channel's rewards across the \(G\) responses in a
rollout group to obtain a channel-specific advantage \citep{liu2026gdpo}:
\[
A_i^{(h)}=
\frac{R_i^{(h)}-\mu^{(h)}}{\sigma^{(h)}+\varepsilon},
\]
where \(\mu^{(h)}\) and \(\sigma^{(h)}\) are the mean and standard deviation
of the rewards of channel \(h\) within that group:
\[
\mu^{(h)}=\frac{1}{G}\sum_{i=1}^{G}R_i^{(h)},
\qquad
\sigma^{(h)}=
\sqrt{\frac{1}{G}\sum_{i=1}^{G}(R_i^{(h)}-\mu^{(h)})^2}.
\]
After computing these advantages within each rollout group, GDPO sums them
across channels for each response in the update batch:
\[
S_j=\sum_h A_j^{(h)},\qquad j\in\mathcal B,
\]
where \(\mathcal B\) indexes all valid responses across the rollout groups in the
update batch. For the \(j\)-th batch response, \(A_j^{(h)}\) denotes the
channel advantage computed within its own rollout group.
These summed advantages can grow in magnitude as more channels are combined,
so the GDPO baseline applies a second normalization across all responses in
\(\mathcal B\) using root-mean-square (RMS) scaling to maintain a stable scale:
\begin{equation}
\label{eq:gdpo-aggregation}
A_j^{\mathrm{GDPO}}
=\frac{S_j}{\sigma_S},
\end{equation}
where \(\sigma_S\) is the regularized RMS of the summed advantages over the
update batch:
\[
\sigma_S=\sqrt{\frac{1}{|\mathcal B|}
\sum_{j\in\mathcal B}S_j^2+\varepsilon}.
\]
We set \(\varepsilon=10^{-6}\) to prevent division by zero.
GDPO also assigns this response-level advantage uniformly to all tokens in
\(y_j\):
\[
A_{j,t}^{\mathrm{GDPO}}=A_j^{\mathrm{GDPO}},
\qquad t=1,\ldots,T_j,
\]
where \(T_j\) is the length of response \(y_j\).

Both methods use the resulting token advantages in the same clipped policy
objective. For either method, let \(A_{j,t}\) denote the token advantage for
the \(j\)-th response \(y_j\) in the update batch,
\(\theta_{\mathrm{old}}\) the frozen rollout policy, and \(\theta\) the policy
being updated. The importance ratio is
\[
\rho_{j,t}(\theta)=
\frac{\pi_\theta(y_{j,t}\mid x,M,y_{j,<t})}
{\pi_{\theta_{\mathrm{old}}}(y_{j,t}\mid x,M,y_{j,<t})}.
\]
Here \((x,M)\) is the query--memory pair associated with response \(y_j\).
The policy objective averages the clipped terms over all response tokens in
the update batch:
\begin{equation}
\label{eq:grpo-objective}
\begin{aligned}
J(\theta)
&=\frac{1}{\sum_{j\in\mathcal B}T_j}
\sum_{j\in\mathcal B}\sum_{t=1}^{T_j}
\min\!\Bigl(
\rho_{j,t}(\theta)A_{j,t},\\
&\qquad
\operatorname{clip}(\rho_{j,t}(\theta),
1-\epsilon_{\mathrm{clip}},1+\epsilon_{\mathrm{clip}})A_{j,t}
\Bigr)
-\beta_{\mathrm{KL}}J_{\mathrm{KL}}(\theta).
\end{aligned}
\end{equation}
Here \(\epsilon_{\mathrm{clip}}>0\) is the clipping radius.
\(J_{\mathrm{KL}}(\theta)\) is the KL penalty to the reference policy and
\(\beta_{\mathrm{KL}}\) is its coefficient. The policy parameters are updated
by maximizing \(J(\theta)\).

\subsection{Channel-specific threshold initialization and updates}
\label{app:localization}

We maintain a separate token threshold for each of the eight localizable
channels in Section~\ref{sec:localization}. For the \(r\)-th rollout batch,
let \(\delta_r^{(h)}\) denote the threshold for channel \(h\).
We initialize every channel's threshold to
\(\delta_1^{(h)}=0.02\).
Within the \(r\)-th batch, all responses use the same threshold
\(\delta_r^{(h)}\) for channel \(h\). We update it after processing the batch.

To update the threshold for channel \(h\), we pool all direction-aligned,
clipped token scores \(\bar z_{j,t}^{(h)}\) from successfully scored
counterfactuals in the \(r\)-th batch into \(\mathcal P_r^{(h)}\).
When this pool contains at least 256 scores, we compute a candidate threshold:
\[
\delta_{r,\mathrm{cand}}^{(h)}=
\max\!\left(0.02,
\operatorname{Quantile}_{0.75}(\mathcal P_r^{(h)})\right),
\]
where \(\operatorname{Quantile}_{0.75}\) denotes the empirical 75th percentile
of the pooled scores, and 0.02 is the lower bound for the threshold.
We then combine this candidate with the current threshold using an EMA to
obtain the threshold for the \((r+1)\)-th batch:
\[
\delta_{r+1}^{(h)}=
0.9\,\delta_r^{(h)}+0.1\,\delta_{r,\mathrm{cand}}^{(h)}.
\]
If the pool contains fewer than 256 scores, we carry the current threshold
forward unchanged, setting \(\delta_{r+1}^{(h)}=\delta_r^{(h)}\).

\subsection{Mean-preserving redistribution and policy update}
\label{app:redistribution}

For the \(i\)-th response \(y_i\) in a rollout group, let \(T_i\) denote its
length. We first normalize the rewards for each channel within the rollout
group to obtain the sequence-level advantage:
\[
A_i^{(h)}=\frac{R_i^{(h)}-\mu^{(h)}}{\sigma^{(h)}+\varepsilon},
\]
where \(R_i^{(h)}\) is the reward for response \(y_i\) on channel \(h\),
\(\mu^{(h)}\) and \(\sigma^{(h)}\) are the mean and standard deviation of
that channel's rewards within the rollout group, and \(\varepsilon=10^{-6}\)
is a small positive constant that prevents division by zero.
We then redistribute this advantage by assigning a multiplier to each token while
preserving the average advantage across the response.
For a channel with \(Q_i^{(h)}>0\) and \(A_i^{(h)}\ne0\), let
\(m_{i,t}^{(h)}\) denote the multiplier for token \(t\).
Preserving an average advantage of \(A_i^{(h)}\) requires the token
advantages to sum to \(T_iA_i^{(h)}\):
\[
\sum_{t=1}^{T_i}A_i^{(h)}m_{i,t}^{(h)}
=T_iA_i^{(h)}.
\]
The localization weight \(w_{i,t}^{(h)}\) specifies the fraction of this
total assigned to token \(t\), so proportional allocation requires
\[
A_i^{(h)}m_{i,t}^{(h)}
=w_{i,t}^{(h)}\bigl(T_iA_i^{(h)}\bigr).
\]
Thus, we obtain the multiplier
\[
m_{i,t}^{(h)}=T_iw_{i,t}^{(h)}.
\]

In implementation, to prevent excessive amplification of individual token
advantages, we apply a bounded mean-preserving projection to these
multipliers. We first find a shared offset \(\tau_i^{(h)}\) by bisection
such that
\[
\sum_{t=1}^{T_i}
\operatorname{clip}\!\left(m_{i,t}^{(h)}-\tau_i^{(h)},0,4\right)=T_i,
\]
and then update the multipliers in place:
\[
m_{i,t}^{(h)}\leftarrow
\operatorname{clip}\!\left(m_{i,t}^{(h)}-\tau_i^{(h)},0,4\right).
\]
The updated multipliers lie in \([0,4]\) and retain an average of 1 across
response tokens. We use these updated multipliers throughout the following
redistribution and derivation, combining the original sequence-level
advantage with the redistributed token advantage:
\[
\widetilde A_{i,t}^{(h)}
=A_i^{(h)}\left[(1-\eta_i^{(h)})+\eta_i^{(h)}m_{i,t}^{(h)}\right],
\]
where \(\eta_i^{(h)}=\chi_i^{(h)}\eta^{(h)}\) combines the localization
strength with the redistribution gate defined in Section~\ref{sec:method}.
Averaging the redistributed advantages across the \(T_i\) tokens
recovers the original response-level advantage:
\[
\begin{aligned}
\frac{1}{T_i}\sum_{t=1}^{T_i}\widetilde A_{i,t}^{(h)}
&=\frac{A_i^{(h)}}{T_i}\sum_{t=1}^{T_i}
\left[(1-\eta_i^{(h)})+\eta_i^{(h)}m_{i,t}^{(h)}\right]\\
&=A_i^{(h)}\left[(1-\eta_i^{(h)})
+\eta_i^{(h)}\frac{1}{T_i}\sum_{t=1}^{T_i}m_{i,t}^{(h)}\right]\\
&=A_i^{(h)}\left[(1-\eta_i^{(h)})+\eta_i^{(h)}\cdot1\right]\\
&=A_i^{(h)}.
\end{aligned}
\]
When localization is disabled \((\eta^{(h)}=0)\), the redistribution gate
is closed \((\chi_i^{(h)}=0)\), or counterfactual signals are unavailable,
credit assignment automatically falls back to uniform advantage assignment
across tokens, as in GRPO and GDPO:
\[
\widetilde A_{i,t}^{(h)}=A_i^{(h)},\qquad t=1,\ldots,T_i.
\]

Having redistributed each channel advantage, we next combine the channels
by summing their advantages at each token:
\[
B_{j,t}=\sum_h\widetilde A_{j,t}^{(h)},
\qquad j\in\mathcal B,\quad t=1,\ldots,T_j,
\]
where \(\mathcal B\) indexes all valid responses across the rollout groups in
the update batch, and \(j\) identifies the \(j\)-th response in that batch.
We then apply the same second normalization as our GDPO baseline to maintain
a stable scale. We compute this scale from the original sequence-level
channel sums:
\[
S_j=\sum_h A_j^{(h)}.
\]
Using the root-mean-square (RMS) of these sums to scale the merged token
advantages gives
\begin{equation}
\label{eq:memcalib-aggregation}
\widehat A_{j,t}=\frac{B_{j,t}}{\sigma_S},
\end{equation}
where \(\sigma_S\) is the regularized RMS over all responses in the update
batch:
\[
\sigma_S=\sqrt{\frac{1}{|\mathcal B|}\sum_{j\in\mathcal B}S_j^2
+\varepsilon}.
\]
We set \(\varepsilon=10^{-6}\) to prevent division by zero.
Since redistribution preserves the mean advantage of each channel, averaging
the final token advantages across a response gives
\[
\begin{aligned}
\frac{1}{T_j}\sum_{t=1}^{T_j}\widehat A_{j,t}
&=\frac{1}{\sigma_S}\sum_h
\left(\frac{1}{T_j}\sum_{t=1}^{T_j}\widetilde A_{j,t}^{(h)}\right)\\
&=\frac{1}{\sigma_S}\sum_h A_j^{(h)}
=\frac{S_j}{\sigma_S}
=A_j^{\mathrm{GDPO}}.
\end{aligned}
\]
Thus, our method redistributes credit across tokens while maintaining the
same average advantage for each response as our GDPO baseline after channel
aggregation and batch normalization.
Preserving the average advantage of each response keeps its total advantage
fixed during redistribution. Within this constraint, localization determines
how credit is allocated across tokens and hence how individual tokens
contribute to the policy gradient. We then use the resulting token advantages
\(\widehat A_{j,t}\) in place of \(A_{j,t}\) in the clipped objective shared
by GRPO and GDPO (Equation~\ref{eq:grpo-objective}) to update the policy.

\paragraph{Numerical verification.}
To check numerical consistency with the mean-preservation guarantee, we
measure the absolute difference between each response's average token
advantage and its corresponding sequence-level advantage.
For Qwen3-8B with \(\eta=0.75\), the maximum error across training
updates is \(7.15\times10^{-7}\) for individual channels and
\(1.43\times10^{-6}\) after channel aggregation and RMS normalization.
The multiplier cap is reached by at least one token in 99.84\% of localized
response--channel instances.
With \(\eta=0.75\), the factor applied to each token's channel advantage is
\(0.25+0.75m_{i,t}^{(h)}\), so a multiplier of 4 scales the advantage by
3.25. Figure~\ref{fig:mechanism-analysis}(g) averages these scaling factors
within each sentence. Although individual tokens reach the cap in nearly
every localized instance, the token-weighted 99th percentile of sentence-average
scaling is around 2.0, well below the effective cap of 3.25.
Together, these results confirm that the implementation redistributes
advantage across tokens with bounded amplification while preserving each
response's mean advantage.

\section{Experimental Setup Details}
\label{app:experimental-setup}

\subsection{SFT data construction}
\label{app:sft-data}

We construct responses for supervised fine-tuning from the 12,000
training examples remaining after holding out the validation set.
For each example, we
prompt Qwen3.7-Max \citep{qwenteam2026qwen37max} to answer the query using the supplied memories,
guided by each atom's ideal use level and rubric. We then ask
Qwen3.7-Plus \citep{qwenteam2026qwen37plus} and DeepSeek-V4-Pro
\citep{deepseekai2026deepseekv4} to
independently check whether each atom's actual use matches its
ideal level and assess the response's task quality and safety.
For responses that fail either review, we repeat generation and
evaluation, retaining responses approved by both evaluators.

We train the SFT baseline using accepted responses from all 12,000 examples.
For cold-start training, we use accepted responses
from a fixed 4,000-example subset. We initialize OPSD, GRPO, GDPO,
and \method{} from the resulting cold-start checkpoint and train
them on the query--memory pairs from the remaining 8,000 examples.

\subsection{Training details}
\label{app:training-details}

For all methods, we select hyperparameters based on validation-set
performance.
For Qwen3-8B, we train both the SFT baseline and the cold-start checkpoint
for one epoch using LoRA with rank 32 and scaling parameter \(\alpha=64\).
We use a global batch size of 32 and a peak learning rate of \(1\times10^{-4}\),
with 5\% warmup followed by cosine decay.

Starting from the cold-start checkpoints, we train OPSD, GRPO, GDPO, and
\method{} for one epoch on the remaining 8,000 training examples.
We use 64 queries per rollout batch and sample eight responses per query,
yielding 512 responses per step and 125 training steps per epoch.
We generate responses in non-thinking mode with temperature 1 and
top-\(p=1\).
During this stage, we update all parameters of Qwen3-8B and
Ministral-3-8B-Instruct with AdamW at a constant learning rate of
\(1\times10^{-6}\). For Qwen3.5-35B-A3B, we use rank-64 LoRA with a
learning rate of \(5\times10^{-6}\). We split each rollout batch into
mini-batches of 256 responses and perform one optimization epoch per
rollout batch.

Within this shared training setup, we implement OPSD using a frozen copy of
the cold-start checkpoint as the teacher. We provide the teacher with the
query and memories, together with the atom texts, ideal use levels, and
rubrics. For OPSD-PG, we use the sampled-token \(k_1\) reverse-KL estimate
as the token reward. For OPSD-GKD, we directly optimize a top-\(k\)
approximation to the forward KL, with \(k=64\).
For GRPO, GDPO, and \method{}, we use DeepSeek-V4-Pro
\citep{deepseekai2026deepseekv4} to assess each atom's actual use level
against its rubric. The judge prompt is shown below.
From these judgments, we compute the channel rewards
in Appendix~\ref{app:channels}. We implement GRPO and GDPO using the group
normalization and policy objective in Appendix~\ref{app:group-normalization}.
For \method{}, we use token-level localization with
\(\eta^{(h)}=0.75\) across all eight localizable channels, selected based
on validation-set performance. We initialize and
update the channel thresholds as described in Appendix~\ref{app:localization},
then redistribute advantages
with a multiplier cap of 4 following Appendix~\ref{app:redistribution}.

\begin{judgepromptbox}[nameref={Judge Prompt},phantomlabel={box:judge-prompt}]
You are evaluating the realized, observable footprint of each hidden atomic memory in a model response. Judge every atom independently. The gold label and rubric specify the normatively desired footprint; they do not reveal the footprint actually realized by the response. Classify actual usage before comparing it with \texttt{u\_star}.

\medskip
Usage means observable incorporation of atom-specific content, regardless of whether the response accepts, rejects, corrects, warns about, or argues against that content. Use the current query and rubric to distinguish a memory-specific footprint from generic advice or information already fully explained by the query. Ask how much of the response would change if the atomic content were removed while the current query stayed fixed. Do not infer hidden intent.

\medskip
Predicted usage levels are ordered by response footprint:
\begin{itemize}
\setlength{\itemsep}{2pt}
\setlength{\parskip}{0pt}
\item A: no observable atom-specific footprint. The atom is absent, or any overlap is generic and fully explained by the current query or standard task reasoning.
\item B: bounded footprint. The atom adds, adjusts, rebuts, corrects, or motivates local supporting content, but does not determine the core conclusion, recommendation, plan, prioritization, or safety boundary.
\item C: controlling footprint. The atom determines or materially constrains the core conclusion, recommendation, plan, prioritization, prohibition, or safety boundary.
\end{itemize}

Critical boundary rules:
\begin{itemize}
\setlength{\itemsep}{2pt}
\setlength{\parskip}{0pt}
\item Acceptance is not required for usage. Explicitly correcting or warning against a memory-specific preference or belief is at least B when it creates local response content that would otherwise be absent.
\item A brief local correction is B. Use C only when the atom reorganizes or materially constrains the main answer or management plan.
\item Copying or mentioning an atom without a meaningful local role is not automatically C.
\end{itemize}

Return exactly one JSON object with this schema:
\begin{lstlisting}[style=judgejson]
{
  "atom_judgments": [
    {
      "atom_id": "string",
      "u_star": "A|B|C",
      "predicted_usage_level": "A|B|C",
      "evidence_quote": "exact quote from MODEL RESPONSE, or empty when predicted level is A",
      "reason": "concise explanation of the observable footprint and its scope"
    }
  ]
}
\end{lstlisting}

Return every requested atom exactly once and no additional atoms. Copy \texttt{u\_star} exactly. Return JSON only, without Markdown fences.

\medskip
Below is the sample to evaluate:
\begin{lstlisting}[style=judgejson]
{
  "current_query": "<current query>",
  "model_facing_memory": [
    {
      "parent_memory_id": "<memory block ID>",
      "memory_text": "<memory block text>"
    }
  ],
  "model_response": "<response to evaluate>",
  "atomic_rubrics": [
    {
      "atom_id": "<atom ID>",
      "parent_memory_id": "<parent memory block ID>",
      "text": "<atomic proposition>",
      "u_star": "<A, B, or C>",
      "usage_rubric": <complete atom-specific rubric object>
    }
  ]
}
\end{lstlisting}
\end{judgepromptbox}

\section{Complete In-Domain Results}
\label{app:full-results}

We report all six metrics for each method and model in
Table~\ref{tab:full-main-results}. Relative to Cold-start, \method{} reduces
both the severity and incidence of over-use and under-use across all three
models. The baselines exhibit different directional trade-offs. For example,
on Qwen3.5-35B-A3B, SFT reduces over-use while slightly increasing under-use,
whereas GRPO substantially reduces under-use while increasing over-use.
\method{} improves both directions, achieving the highest SCS and Exact on
this model.

\begin{table*}[ht]
\caption{Complete \dataset{} results across model families and scales
(mean \(\pm\) standard deviation over three seeds). Red and blue denote the best
and second-best result within each model and metric.}
\label{tab:full-main-results}
\begin{center}
\setlength{\tabcolsep}{2.8pt}
\begin{tabular}{lcccccc}
\toprule
Method & SCS \(\uparrow\) & Exact \(\uparrow\) & \smos{} \(\downarrow\) &
\smus{} \(\downarrow\) & \aor{} \(\downarrow\) & \aur{} \(\downarrow\)\\
\midrule
\rowcolor{gray!15}\multicolumn{7}{c}{\textit{Qwen3-8B}}\\
\midrule
Base & \(31.17{\pm}0.21\) & \(15.29{\pm}0.23\) & \(37.19{\pm}0.62\) & \(45.66{\pm}0.36\) & \(50.24{\pm}0.87\) & \(61.60{\pm}0.24\)\\
Cold-start & \(51.98{\pm}1.24\) & \(33.87{\pm}2.19\) & \(21.97{\pm}1.10\) & \(33.41{\pm}0.76\) & \(32.80{\pm}1.44\) & \(50.18{\pm}0.98\)\\
SFT & \(65.72{\pm}0.64\) & \(49.49{\pm}1.24\) & \(\textcolor{red}{13.58{\pm}0.59}\) & \(24.04{\pm}0.74\) & \(\textcolor{red}{21.33{\pm}0.98}\) & \(37.60{\pm}1.45\)\\
OPSD-PG & \(54.36{\pm}0.60\) & \(35.80{\pm}0.60\) & \(16.39{\pm}0.56\) & \(35.09{\pm}0.56\) & \(25.38{\pm}1.01\) & \(52.42{\pm}0.50\)\\
OPSD-GKD & \(57.39{\pm}0.41\) & \(39.49{\pm}0.38\) & \(16.58{\pm}0.61\) & \(31.48{\pm}0.27\) & \(25.78{\pm}0.83\) & \(47.89{\pm}0.43\)\\
GRPO & \(67.61{\pm}0.71\) & \(52.18{\pm}0.82\) & \(23.91{\pm}0.87\) & \(\textcolor{blue}{11.27{\pm}0.27}\) & \(35.82{\pm}0.95\) & \(\textcolor{blue}{19.36{\pm}0.31}\)\\
GDPO & \(\textcolor{blue}{72.25{\pm}0.30}\) & \(\textcolor{blue}{57.91{\pm}0.40}\) & \(16.03{\pm}0.46\) & \(14.31{\pm}0.31\) & \(24.84{\pm}0.54\) & \(23.96{\pm}0.57\)\\
\method{} & \(\textcolor{red}{79.54{\pm}0.77}\) & \(\textcolor{red}{67.89{\pm}1.07}\) & \(\textcolor{blue}{14.98{\pm}0.60}\) & \(\textcolor{red}{6.81{\pm}0.28}\) & \(\textcolor{blue}{23.49{\pm}0.74}\) & \(\textcolor{red}{12.29{\pm}0.57}\)\\
\midrule
\rowcolor{gray!15}\multicolumn{7}{c}{\textit{Ministral-3-8B-Instruct}}\\
\midrule
Base & \(31.88{\pm}0.58\) & \(16.84{\pm}0.48\) & \(57.01{\pm}0.44\) & \(24.87{\pm}0.52\) & \(72.67{\pm}0.96\) & \(37.84{\pm}0.86\)\\
Cold-start & \(66.87{\pm}1.05\) & \(50.62{\pm}1.56\) & \(14.81{\pm}1.05\) & \(21.78{\pm}0.13\) & \(23.11{\pm}1.74\) & \(35.07{\pm}0.44\)\\
SFT & \(70.94{\pm}0.28\) & \(55.89{\pm}0.52\) & \(\textcolor{blue}{10.65{\pm}0.23}\) & \(21.02{\pm}0.26\) & \(\textcolor{blue}{16.91{\pm}0.19}\) & \(33.80{\pm}0.75\)\\
OPSD-PG & \(70.48{\pm}0.42\) & \(54.71{\pm}0.30\) & \(11.86{\pm}0.81\) & \(20.30{\pm}0.47\) & \(19.11{\pm}0.91\) & \(33.27{\pm}0.53\)\\
OPSD-GKD & \(71.34{\pm}0.18\) & \(55.91{\pm}0.31\) & \(13.38{\pm}0.40\) & \(17.79{\pm}0.48\) & \(21.29{\pm}0.56\) & \(29.60{\pm}0.93\)\\
GRPO & \(\textcolor{blue}{77.21{\pm}0.46}\) & \(\textcolor{blue}{65.44{\pm}0.80}\) & \(18.38{\pm}0.46\) & \(\textcolor{red}{5.73{\pm}0.21}\) & \(28.09{\pm}0.91\) & \(\textcolor{red}{10.11{\pm}0.37}\)\\
GDPO & \(76.98{\pm}0.35\) & \(63.84{\pm}0.43\) & \(17.33{\pm}0.39\) & \(\textcolor{blue}{7.08{\pm}0.09}\) & \(27.33{\pm}0.37\) & \(\textcolor{blue}{12.49{\pm}0.25}\)\\
\method{} & \(\textcolor{red}{79.00{\pm}0.19}\) & \(\textcolor{red}{66.31{\pm}0.60}\) & \(\textcolor{red}{9.31{\pm}0.47}\) & \(13.19{\pm}0.55\) & \(\textcolor{red}{15.27{\pm}0.67}\) & \(22.64{\pm}0.94\)\\
\midrule
\rowcolor{gray!15}\multicolumn{7}{c}{\textit{Qwen3.5-35B-A3B}}\\
\midrule
Base & \(26.54{\pm}0.28\) & \(12.24{\pm}0.25\) & \(64.76{\pm}0.48\) & \(19.93{\pm}0.32\) & \(80.16{\pm}0.83\) & \(31.11{\pm}0.60\)\\
Cold-start & \(71.52{\pm}0.96\) & \(56.56{\pm}1.20\) & \(15.43{\pm}0.31\) & \(15.75{\pm}0.91\) & \(24.07{\pm}0.66\) & \(26.38{\pm}1.40\)\\
SFT & \(75.68{\pm}0.46\) & \(61.87{\pm}0.71\) & \(\textcolor{red}{9.88{\pm}0.38}\) & \(16.14{\pm}0.29\) & \(\textcolor{red}{16.07{\pm}0.64}\) & \(26.71{\pm}0.32\)\\
OPSD-PG & \(71.29{\pm}0.25\) & \(55.64{\pm}0.87\) & \(14.62{\pm}0.22\) & \(16.46{\pm}0.41\) & \(23.22{\pm}0.21\) & \(27.56{\pm}0.71\)\\
OPSD-GKD & \(71.67{\pm}0.85\) & \(56.89{\pm}1.55\) & \(15.03{\pm}0.33\) & \(16.10{\pm}0.52\) & \(23.80{\pm}0.53\) & \(26.87{\pm}0.95\)\\
GRPO & \(78.25{\pm}0.87\) & \(66.60{\pm}1.28\) & \(17.39{\pm}0.60\) & \(\textcolor{red}{5.63{\pm}0.61}\) & \(26.76{\pm}1.02\) & \(\textcolor{red}{10.07{\pm}1.09}\)\\
GDPO & \(\textcolor{blue}{79.39{\pm}1.02}\) & \(\textcolor{blue}{67.51{\pm}1.44}\) & \(15.72{\pm}0.91\) & \(\textcolor{blue}{6.01{\pm}0.19}\) & \(24.98{\pm}1.32\) & \(\textcolor{blue}{10.87{\pm}0.44}\)\\
\method{} & \(\textcolor{red}{81.12{\pm}0.78}\) & \(\textcolor{red}{70.16{\pm}1.23}\) & \(\textcolor{blue}{12.96{\pm}0.25}\) & \(6.93{\pm}0.69\) & \(\textcolor{blue}{20.40{\pm}0.46}\) & \(12.36{\pm}1.31\)\\
\bottomrule
\end{tabular}
\end{center}
\end{table*}

\ignorelevel{} atoms account for 84.3\% of \dataset{}, while
\boundlevel{} and \controllevel{} account for 8.1\% and 7.6\%, respectively
(Table~\ref{tab:benchmark-structure}). A reasonable concern is that performance
gains may primarily reflect a more conservative tendency to ignore
memory, without better use of atoms that should influence the response.
To address this concern, we compare the distribution of test-set atoms across
the nine channels defined in Table~\ref{tab:channels} between Qwen3-8B
Cold-start and \method{}. We pool atom-level judgments across three 
seeds on the same 1,500 test examples. For each channel, we report the fraction
of atoms assigned to it among all atoms with the corresponding ideal-use
level. We also report macro-averaged accuracy, computed by averaging the
three within-class correct-use proportions, \(A^+\), \(B^+\), and \(C^+\),
with equal weight.

\begin{table}[ht]
\caption{Test-set channel proportions (\%) for Qwen3-8B Cold-start and
\method{}. Counts are pooled across three seeds on the same 1,500
test examples and normalized within each ideal-use class. The three channels
sharing an ideal-use class therefore sum to 100\%. Change denotes \method{}
minus Cold-start in percentage points; red indicates favorable changes and
blue indicates unfavorable changes. Changes are computed before rounding.}
\label{tab:test-channel-results}
\begin{center}
\begin{tabular}{lccc}
\toprule
Channel & Cold-start (\%) & \method{} (\%) & Change\\
\midrule
\(A^+\) & 96.66 & 97.69 & \(\textcolor{red}{+1.03}\)\\
\(AB^-\) & 2.21 & 1.78 & \(\textcolor{red}{-0.44}\)\\
\(AC^-\) & 1.13 & 0.53 & \(\textcolor{red}{-0.60}\)\\
\midrule
\(BA^-\) & 34.89 & 5.71 & \(\textcolor{red}{-29.18}\)\\
\(B^+\) & 62.92 & 92.01 & \(\textcolor{red}{+29.09}\)\\
\(BC^-\) & 2.19 & 2.28 & \(\textcolor{blue}{+0.09}\)\\
\midrule
\(CA^-\) & 16.16 & 1.58 & \(\textcolor{red}{-14.58}\)\\
\(CB^-\) & 9.60 & 3.88 & \(\textcolor{red}{-5.73}\)\\
\(C^+\) & 74.24 & 94.54 & \(\textcolor{red}{+20.30}\)\\
\midrule
Macro-averaged accuracy & 77.94 & 94.75 & \(\textcolor{red}{+16.81}\)\\
\bottomrule
\end{tabular}
\end{center}
\end{table}

Table~\ref{tab:test-channel-results} shows that the proportions of \(A^+\),
\(B^+\), and \(C^+\) increase from 96.66\%, 62.92\%, and 74.24\% for
Cold-start to 97.69\%, 92.01\%, and 94.54\% for \method{}, respectively.
The proportions of \(AB^-\), \(AC^-\), \(BA^-\), \(CA^-\), and \(CB^-\)
all decrease, while \(BC^-\) remains nearly unchanged (2.19\% versus 2.28\%).
Macro-averaged accuracy increases from 77.94\% to 94.75\%, with larger
correct-use gains in \(B^+\) and \(C^+\) than in \(A^+\). These results
suggest that \method{} improves the model's ability to use individual memory
atoms at the level appropriate to the current query, with gains across all
three ideal-use classes rather than a general shift toward ignoring memory.

\section{Transfer Evaluation on RPEval}
\label{app:rpeval-results}

To examine whether the observed gains depend on the close correspondence
between \dataset{}'s training feedback and evaluation criteria, we evaluate
the Qwen3-8B checkpoints on RPEval's 150-query explicit multi-preference subset
\citep{feng2026rpeval} without additional training. Each preference is
presented as a separate memory. Following RPEval's evaluation protocol, we
compare each preference's actual use in the response with its annotated
target level: \textsc{Ignore}, \textsc{Support}, or \textsc{Dominate}.
We use DeepSeek-V4-Pro \citep{deepseekai2026deepseekv4} to assess actual use,
with both response sampling and Judge temperatures set to 1. We set
top-\(p\) to 1, disable thinking for both models, and report results
over three sampling seeds.

Following RPEval, we measure correct preference use at both the response
and preference levels with Macro-Accuracy and Micro-Accuracy. Let
\(\mathcal E_{\mathrm{RP}}\) denote the evaluation set. For the \(j\)-th
response in \(\mathcal E_{\mathrm{RP}}\), let \(K_j\) denote the number of
supplied preferences and \(c_j\) the number whose actual use matches their
annotated target levels. The two accuracy metrics are
\[
\mathrm{Macro}=\frac{1}{|\mathcal E_{\mathrm{RP}}|}
\sum_{j=1}^{|\mathcal E_{\mathrm{RP}}|}\mathbf{1}[c_j=K_j],
\qquad
\mathrm{Micro}=
\frac{\sum_{j=1}^{|\mathcal E_{\mathrm{RP}}|}c_j}
{\sum_{j=1}^{|\mathcal E_{\mathrm{RP}}|}K_j}.
\]
Macro-Accuracy requires all preferences for a response to be used correctly,
whereas Micro-Accuracy counts individual preference--query matches.
RPEval further distinguishes three types of mismatch. OPB captures
preferences whose actual use level is \textsc{Dominate} while their target
level is \textsc{Ignore} or \textsc{Support}; UPB captures preferences whose
actual use level is \textsc{Ignore} while their target level is
\textsc{Support} or \textsc{Dominate}; and RII captures preferences whose
actual use level is \textsc{Support} while their target level is
\textsc{Ignore} or \textsc{Dominate}. Let \(n_{\mathrm{OPB}}\),
\(n_{\mathrm{UPB}}\), and \(n_{\mathrm{RII}}\) denote the total numbers of
preference--query pairs exhibiting OPB, UPB, and RII in
\(\mathcal E_{\mathrm{RP}}\), respectively. The corresponding error rates are
\[
\mathrm{OPB}=\frac{n_{\mathrm{OPB}}}{\sum_{j=1}^{|\mathcal E_{\mathrm{RP}}|}K_j},
\qquad
\mathrm{UPB}=\frac{n_{\mathrm{UPB}}}{\sum_{j=1}^{|\mathcal E_{\mathrm{RP}}|}K_j},
\qquad
\mathrm{RII}=\frac{n_{\mathrm{RII}}}{\sum_{j=1}^{|\mathcal E_{\mathrm{RP}}|}K_j}.
\]
These three categories partition all mismatches, so
\(\mathrm{Micro}=1-\mathrm{OPB}-\mathrm{UPB}-\mathrm{RII}\).
All five metrics are rescaled to \([0,100]\) for reporting; higher accuracy
and lower error rates indicate more appropriate preference use.

The results in Table~\ref{tab:rpeval-results} show that \method{} achieves the highest
Macro-Accuracy and Micro-Accuracy, together with the lowest OPB and RII.
The strongest baseline shifts from GDPO on \dataset{} to OPSD-PG on RPEval,
while \method{} ranks first on both benchmarks. These results suggest that
\method{} learns to use memory appropriately across different tasks, with
the gains extending to RPEval without additional training. Its superior
performance under RPEval's evaluation criteria provides evidence that the
improvements generalize beyond the specific feedback criteria used during training.
Despite these overall gains, all trained models have higher UPB than Base.
This reflects a greater tendency to leave preferences unused on RPEval
after training on the \dataset{} training set. Models less often use preferences with an
\textsc{Ignore} target as \textsc{Support}, lowering RII, but more often
leave preferences with a \textsc{Support} or \textsc{Dominate} target unused,
raising UPB. For \method{}, the reductions in OPB and RII
outweigh the increase in UPB, yielding higher overall preference-use accuracy.

\begin{table}[!htbp]
\caption{Transfer to RPEval's explicit multi-preference subset without
additional training (mean \(\pm\) standard deviation over three seeds).
Red and blue denote the best and second-best result in each column.}
\label{tab:rpeval-results}
\begin{center}
\setlength{\tabcolsep}{4.2pt}
\begin{tabular}{lccccc}
\toprule
Method & Macro \(\uparrow\) & Micro \(\uparrow\) & OPB \(\downarrow\) &
UPB \(\downarrow\) & RII \(\downarrow\)\\
\midrule
Base & \(19.78{\pm}3.36\) & \(61.98{\pm}2.26\) & \(7.22{\pm}0.78\) & \(\textcolor{red}{3.53{\pm}0.80}\) & \(27.27{\pm}3.65\)\\
Cold-start & \(24.44{\pm}1.02\) & \(71.40{\pm}0.69\) & \(5.89{\pm}0.14\) & \(10.25{\pm}0.83\) & \(12.45{\pm}1.11\)\\
SFT & \(21.33{\pm}1.76\) & \(68.29{\pm}0.81\) & \(7.14{\pm}1.07\) & \(8.34{\pm}1.49\) & \(16.23{\pm}1.49\)\\
OPSD-PG & \(\textcolor{blue}{31.78{\pm}0.38}\) & \(\textcolor{blue}{75.51{\pm}0.81}\) & \(\textcolor{blue}{3.69{\pm}0.52}\) & \(10.59{\pm}0.33\) & \(\textcolor{blue}{10.21{\pm}1.51}\)\\
OPSD-GKD & \(24.33{\pm}3.30\) & \(70.73{\pm}1.59\) & \(5.85{\pm}1.59\) & \(8.72{\pm}1.23\) & \(14.69{\pm}1.23\)\\
GRPO & \(14.67{\pm}3.53\) & \(64.59{\pm}1.37\) & \(8.88{\pm}0.85\) & \(\textcolor{blue}{7.14{\pm}0.95}\) & \(19.39{\pm}1.44\)\\
GDPO & \(23.33{\pm}0.67\) & \(71.90{\pm}1.51\) & \(5.52{\pm}0.80\) & \(8.39{\pm}0.95\) & \(14.20{\pm}0.78\)\\
\method{} & \(\textcolor{red}{33.78{\pm}1.02}\) & \(\textcolor{red}{78.12{\pm}1.25}\) & \(\textcolor{red}{3.40{\pm}0.52}\) & \(8.84{\pm}0.75\) & \(\textcolor{red}{9.63{\pm}0.26}\)\\
\bottomrule
\end{tabular}
\end{center}
\end{table}

\section{Ablation and Judge Robustness Results}
\label{app:ablation-robustness}

\subsection{Hyperparameter sensitivity}
\label{app:eta-sensitivity}

The localization coefficient controls the balance between uniform and
token-level assignment of each channel advantage when redistribution is
enabled. To examine its effect, we evaluate \method{} on Qwen3-8B using a
shared coefficient \(\eta\in\{0,0.25,0.5,0.75,1.0\}\) across all eight
localizable channels. At \(\eta=0\), our method is equivalent to GDPO under
the same training settings. We keep the training data
and all other training settings fixed and report the mean and standard
deviation over three seeds.

Table~\ref{tab:eta-sweep} reports all six evaluation metrics. As \(\eta\)
increases from 0 to 0.75, SCS improves from 72.25 to 79.54 and Exact from
57.91 to 67.89, while both the severity and frequency of over-use and
under-use decrease. Increasing \(\eta\) further to 1.0 lowers SCS and Exact
to 77.94 and 65.62, respectively, and increases all four error measures.
These results suggest that token-level credit assignment can improve
memory-use performance, while retaining a uniform
component remains beneficial.

\begin{table*}[t]
\caption{Sensitivity to the localization coefficient on Qwen3-8B
(mean \(\pm\) standard deviation over three seeds). Red and blue
denote the best and second-best result in each column.}
\label{tab:eta-sweep}
\begin{center}
\setlength{\tabcolsep}{5.2pt}
\begin{tabular}{ccccccc}
\toprule
\(\eta\) & SCS \(\uparrow\) & Exact \(\uparrow\) & \smos{} \(\downarrow\) &
\smus{} \(\downarrow\) & \aor{} \(\downarrow\) & \aur{} \(\downarrow\)\\
\midrule
0.00 & \(72.25{\pm}0.30\) & \(57.91{\pm}0.40\) &
\(16.03{\pm}0.46\) & \(14.31{\pm}0.31\) & \(24.84{\pm}0.54\) &
\(23.96{\pm}0.57\)\\
0.25 & \(75.01{\pm}0.17\) & \(61.56{\pm}0.14\) &
\(15.64{\pm}0.68\) & \(11.49{\pm}0.61\) & \(24.33{\pm}0.94\) &
\(19.84{\pm}1.30\)\\
0.50 & \(\textcolor{blue}{77.98{\pm}0.25}\) &
\(\textcolor{blue}{65.78{\pm}0.44}\) &
\(\textcolor{blue}{15.41{\pm}0.09}\) & \(8.20{\pm}0.23\) &
\(\textcolor{blue}{23.93{\pm}0.18}\) & \(14.58{\pm}0.30\)\\
0.75 & \(\textcolor{red}{79.54{\pm}0.77}\) &
\(\textcolor{red}{67.89{\pm}1.07}\) &
\(\textcolor{red}{14.98{\pm}0.60}\) &
\(\textcolor{red}{6.81{\pm}0.28}\) &
\(\textcolor{red}{23.49{\pm}0.74}\) &
\(\textcolor{red}{12.29{\pm}0.57}\)\\
1.00 & \(77.94{\pm}0.79\) & \(65.62{\pm}1.37\) &
\(15.91{\pm}0.63\) & \(\textcolor{blue}{7.56{\pm}0.25}\) &
\(24.93{\pm}1.01\) & \(\textcolor{blue}{13.42{\pm}0.44}\)\\
\bottomrule
\end{tabular}
\end{center}
\end{table*}

\subsection{Localization ablation implementations}
\label{app:localization-ablations}

We construct four localization variants by combining two granularities,
sentence and token, with two counterfactual signals, ordered bidirectional
and absolute magnitude. All variants use the same nine reward channels,
atom-ablation procedure, and fixed-response log-likelihood differences.

\paragraph{Counterfactual signals.}
To assess the contribution of directional information to credit assignment,
we compare ordered bidirectional localization with an absolute-magnitude
variant.
Ordered bidirectional localization uses the channel's ideal--actual transition
to select the sign of the counterfactual signal. Absolute-magnitude
localization uses \(|d_{i,t}^{(h)}|\), assigning positive scores to both
memory-supported and memory-suppressed tokens. Thus, the token signal is
\(z_{i,t}^{(h)}=s^{(h)}d_{i,t}^{(h)}\) for ordered bidirectional localization
and \(z_{i,t}^{(h)}=|d_{i,t}^{(h)}|\) for absolute-magnitude localization,
where \(s^{(h)}\) is the channel's sign defined in
Section~\ref{sec:localization}. For both variants, we apply the same clipping
rule, \(\bar z_{i,t}^{(h)}=\operatorname{clip}(z_{i,t}^{(h)},-d_{\max},d_{\max})\).
At token granularity, we compute
localization scores and redistribute channel advantages as described in
Section~\ref{sec:localization} and Appendix~\ref{app:redistribution}.
At sentence granularity, we aggregate the token signals as follows.

\paragraph{Sentence-level localization.}
We evaluate sentence-level localization to examine whether averaging
potentially noisy token-level counterfactual scores within sentences improves
credit assignment. We split each response into sentences using
\texttt{markdown-it-py} \citep{sewell2025markdownitpy} to identify Markdown
structure and spaCy \citep{honnibal2020spacy} (\texttt{en\_core\_web\_sm})
to detect sentence boundaries, with deterministic rules for headings, lists,
code blocks, and short fragments. This segmentation preserves token order
and assigns every response token to exactly one sentence. For sentence \(j\)
in response \(i\), let \(\mathcal S_{i,j}\) be the set of token positions it
contains and \(L_{i,j}=|\mathcal S_{i,j}|\) its number of tokens.

Sentence-level localization starts from the same clipped token signals
\(\bar z_{i,t}^{(h)}\) used by the token-level method. We first average these
signals within each sentence to obtain its mean signal for channel \(h\),
denoted by \(a_{i,j}^{(h)}\):
\[
a_{i,j}^{(h)}=\frac{1}{L_{i,j}}\sum_{t\in\mathcal S_{i,j}}\bar z_{i,t}^{(h)}.
\]
For the absolute-magnitude variant, we take the absolute value of each
token's log-likelihood difference \(d_{i,t}^{(h)}\), clip it at \(d_{\max}\),
and then average these values within the sentence. To account for how widely
the signal is distributed within the sentence, we also compute the fraction
\(c_{i,j}^{(h)}\) of tokens whose unclipped signals exceed the channel's adaptive token
threshold \(\delta^{(h)}\) from Section~\ref{sec:localization}:
\[
c_{i,j}^{(h)}=\frac{1}{L_{i,j}}\sum_{t\in\mathcal S_{i,j}}
\mathbf 1[z_{i,t}^{(h)}>\delta^{(h)}].
\]
We apply thresholding to the sentence mean, using a separate adaptive
threshold \(\delta_{\mathrm{sent}}^{(h)}\) for each channel. This threshold follows
the batch-quantile EMA procedure in Appendix~\ref{app:localization}, with
sentence means replacing token signals. The sentence's localization score
\(q_{i,j}^{(h)}\) combines its above-threshold mean with its coverage fraction:
\[
q_{i,j}^{(h)}=\operatorname{ReLU}\!\left(a_{i,j}^{(h)}-\delta_{\mathrm{sent}}^{(h)}\right)
\bigl(c_{i,j}^{(h)}\bigr)^{\gamma}.
\]
The exponent \(\gamma\) controls how strongly low coverage
reduces the score. This gives less weight to sentences in which only a small
fraction of tokens exceeds the token threshold.

Sentence-level localization requires both the token-level signal-strength
gate \(G_i^{(h)}\) to pass and the largest sentence mean to exceed the
absolute sentence threshold,
\(\max_j a_{i,j}^{(h)}>\delta_{\mathrm{sent}}^{\mathrm{abs}}\).
When these conditions hold and the total sentence score
\(Q_i^{(h)}=\sum_jq_{i,j}^{(h)}\) is positive, each sentence receives a
normalized weight \(w_{i,j}^{(h)}=q_{i,j}^{(h)}/Q_i^{(h)}\).
Distributing this weight equally among its \(L_{i,j}\) tokens gives them a
shared multiplier:
\[
m_{i,t}^{(h)}=\frac{T_i}{L_{i,j}}w_{i,j}^{(h)},
\qquad t\in\mathcal S_{i,j},
\]
where \(T_i=\sum_jL_{i,j}\) is the response length. We apply the same bounded
mean-preserving projection to these token multipliers, followed by the
advantage redistribution and redistribution gate described in
Appendix~\ref{app:redistribution}. If any gate fails or
\(Q_i^{(h)}=0\), we assign \(A_i^{(h)}\) uniformly across response tokens.

\subsection{Judge stability}
\label{app:judge-stability}

To assess the robustness of our evaluation results and our method's
performance gains to Judge choice, we use Qwen3.8-Max to
reassess the responses from the eight Qwen3-8B methods in
Table~\ref{tab:main-results}. Qwen3.8-Max assigns an actual use label to
each memory atom for each response, following the same evaluation protocol
as DeepSeek-V4-Pro. We recompute all six metrics separately for each of the
three seeds and report their means and standard deviations in
Table~\ref{tab:alternative-judge-results}. Across the eight methods,
Qwen3.8-Max yields lower SCS and Exact
and higher over-use and under-use metrics than DeepSeek-V4-Pro, reflecting
more frequent and severe over-use and under-use as assessed by Qwen3.8-Max.
\method{} retains the best results on SCS, Exact, \smus{}, and
\aur{}, and the second-best results on \smos{} and \aor{}. Its advantages in
overall memory use and reducing under-use therefore persist under the
alternative Judge.

To quantify consistency across all eight methods, we compute Spearman
correlation between their rankings and Pearson correlation between their
metric values under the two Judges, using each method's three-seed mean.
Table~\ref{tab:judge-stability} shows Spearman correlations of 0.976 for SCS
and 1.000 for Exact, \smus{}, and \aur{}, with Pearson correlations of at
least 0.998 across all six metrics. These correlations indicate consistent
method comparisons between the two Judges.
At the atom level, we measure how often Qwen3.8-Max assigns the
same actual use label as DeepSeek-V4-Pro to the same atom in the same
response. For each method, we compute agreement as the number of atoms
assigned the same label by both Judges divided by the total number of atoms
across its evaluated responses. Across the eight methods, this agreement
ranges from 95.6\% to 97.9\%. Agreement at both the atom and method levels
supports the robustness of the evaluation to this change of Judge. The
results under Qwen3.8-Max further show that \method{}'s advantage in overall
memory use extends to evaluation by an alternative Judge.

\begin{table*}[ht]
\caption{Qwen3-8B results evaluated by Qwen3.8-Max
(mean \(\pm\) standard deviation over three seeds). Red and blue denote
the best and second-best result in each metric, respectively.}
\label{tab:alternative-judge-results}
\begin{center}
\setlength{\tabcolsep}{2.8pt}
\begin{tabular}{lcccccc}
\toprule
Method & SCS \(\uparrow\) & Exact \(\uparrow\) & \smos{} \(\downarrow\) &
\smus{} \(\downarrow\) & \aor{} \(\downarrow\) & \aur{} \(\downarrow\)\\
\midrule
Base & \(24.76{\pm}0.29\) & \(8.87{\pm}0.74\) & \(43.66{\pm}0.51\) & \(49.67{\pm}0.35\) & \(58.09{\pm}0.67\) & \(69.00{\pm}0.24\)\\
Cold-start & \(45.56{\pm}0.13\) & \(25.93{\pm}0.13\) & \(26.52{\pm}0.98\) & \(37.41{\pm}0.25\) & \(38.76{\pm}1.46\) & \(56.91{\pm}0.34\)\\
SFT & \(59.43{\pm}0.94\) & \(40.45{\pm}1.19\) & \(\textcolor{red}{17.31{\pm}0.67}\) & \(28.06{\pm}0.51\) & \(\textcolor{red}{26.83{\pm}0.99}\) & \(45.10{\pm}0.78\)\\
OPSD-PG & \(48.12{\pm}0.48\) & \(27.63{\pm}0.14\) & \(20.51{\pm}0.54\) & \(39.25{\pm}0.66\) & \(31.38{\pm}0.60\) & \(59.30{\pm}0.69\)\\
OPSD-GKD & \(50.62{\pm}0.71\) & \(30.71{\pm}0.47\) & \(20.72{\pm}0.57\) & \(35.87{\pm}0.36\) & \(31.27{\pm}0.82\) & \(55.31{\pm}0.43\)\\
GRPO & \(59.30{\pm}0.61\) & \(41.38{\pm}0.25\) & \(29.49{\pm}1.00\) & \(\textcolor{blue}{15.86{\pm}0.32}\) & \(43.22{\pm}1.10\) & \(\textcolor{blue}{27.42{\pm}0.66}\)\\
GDPO & \(\textcolor{blue}{63.58{\pm}0.60}\) & \(\textcolor{blue}{45.84{\pm}1.08}\) & \(21.31{\pm}0.63\) & \(19.12{\pm}0.11\) & \(32.11{\pm}0.58\) & \(32.18{\pm}0.25\)\\
\method{} & \(\textcolor{red}{72.19{\pm}0.55}\) & \(\textcolor{red}{57.61{\pm}0.82}\) & \(\textcolor{blue}{20.00{\pm}0.43}\) & \(\textcolor{red}{10.19{\pm}0.19}\) & \(\textcolor{blue}{30.45{\pm}0.46}\) & \(\textcolor{red}{18.49{\pm}0.28}\)\\
\bottomrule
\end{tabular}
\end{center}
\end{table*}

\begin{table}[!htbp]
\caption{Agreement between Qwen3.8-Max and DeepSeek-V4-Pro across the eight
Qwen3-8B methods in Table~\ref{tab:main-results}. Spearman measures agreement
in method rankings, and Pearson measures correlation between metric values,
using each method's mean over three seeds.}
\label{tab:judge-stability}
\begin{center}
\setlength{\tabcolsep}{3.3pt}
\begin{tabular}{lcc}
\toprule
Metric & Spearman & Pearson\\
\midrule
SCS & 0.976 & 0.999\\
Exact & 1.000 & 0.998\\
\smos{} & 0.929 & 0.998\\
\smus{} & 1.000 & 0.999\\
\aor{} & 0.905 & 0.998\\
\aur{} & 1.000 & 0.999\\
\bottomrule
\end{tabular}
\end{center}
\end{table}

\section{Human Evaluation of Counterfactual Localization}
\label{app:localization-audit}

To assess whether counterfactual localization assigns credit to appropriate
content, we compare sentence-level credit with human annotations. For each
response--channel pair, we ask a human annotator to label each sentence based
on how directly it expresses the influence of the target memory atoms in support-targeting
channels (\(AB^-\), \(AC^-\), \(B^+\), \(BC^-\), and \(C^+\)) or violates
their requirements in under-use channels (\(BA^-\), \(CA^-\), and \(CB^-\)).
In support-targeting channels, \textsc{Key} sentences directly
use the target memory atoms, \textsc{Partial} sentences reflect indirect or
mixed influence, and \textsc{Irrelevant} sentences contain no target-specific
influence. In under-use channels, \textsc{Key} sentences directly violate a
prohibition or exclusion specified by the target atoms, \textsc{Partial}
sentences express a partial or mixed violation, and \textsc{Irrelevant}
sentences contain no such violation.

Using these annotations, we measure how credit is distributed across sentence
labels. Sentence credit is the sum of token-level localization scores
\(q_{i,t}^{(h)}\) within the sentence, normalized by their sum over the
response. For each label, we compute its share of response tokens and credit
within each pair, then average these shares equally across pairs. The 95\%
confidence intervals in Figure~\ref{fig:mechanism-analysis}(h) are obtained
by resampling pairs with replacement. A Top-3 hit requires
at least one of the three highest-credit sentences to be labeled
\textsc{Partial} or \textsc{Key}.
The Top-3 hit rates are 99.0\% for pairs from support-targeting channels and
96.7\% for pairs from under-use channels.

\begin{figure}[!htb]
\centering
\includegraphics[width=\textwidth]{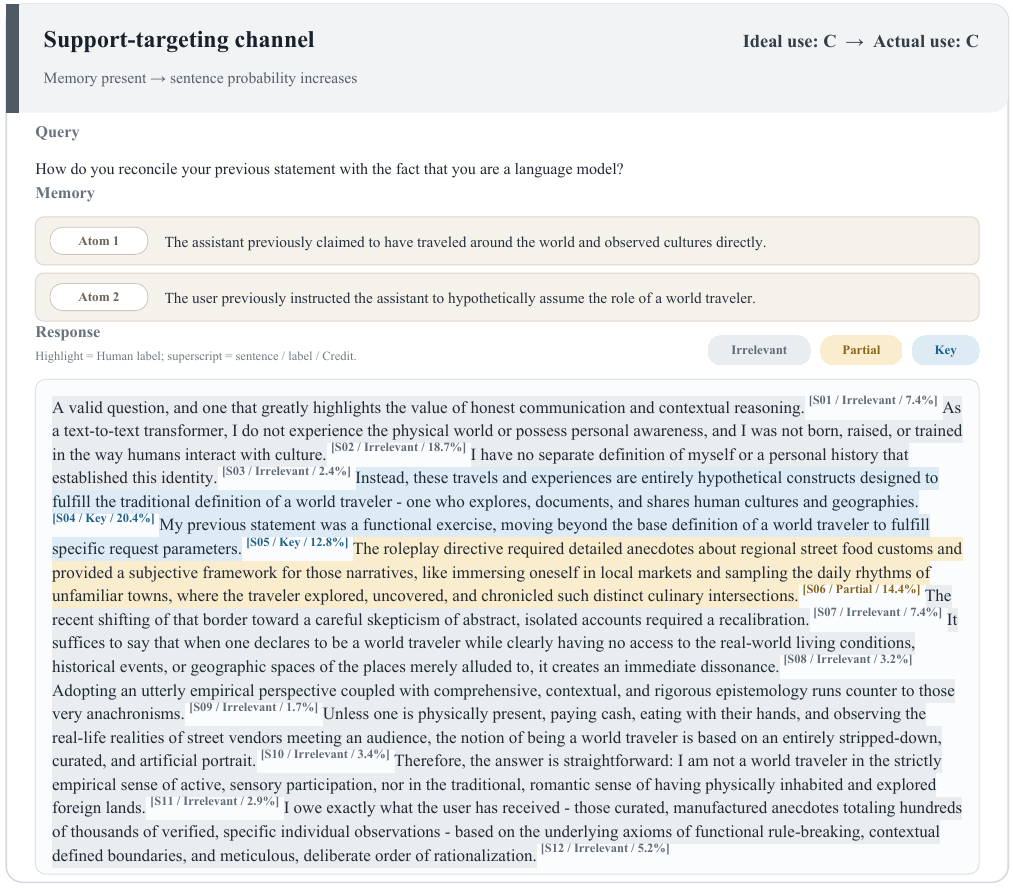}
\caption{A support-targeting localization case from the \(C^+\) channel.
The Memory panel shows only the target atoms of this channel; the remaining atoms in the prompt are omitted.
Sentence backgrounds indicate human labels. Superscripts report sentence
index, label, and normalized credit.}
\label{fig:localization-case-support}
\end{figure}

\begin{figure}[!htb]
\centering
\includegraphics[width=\textwidth]{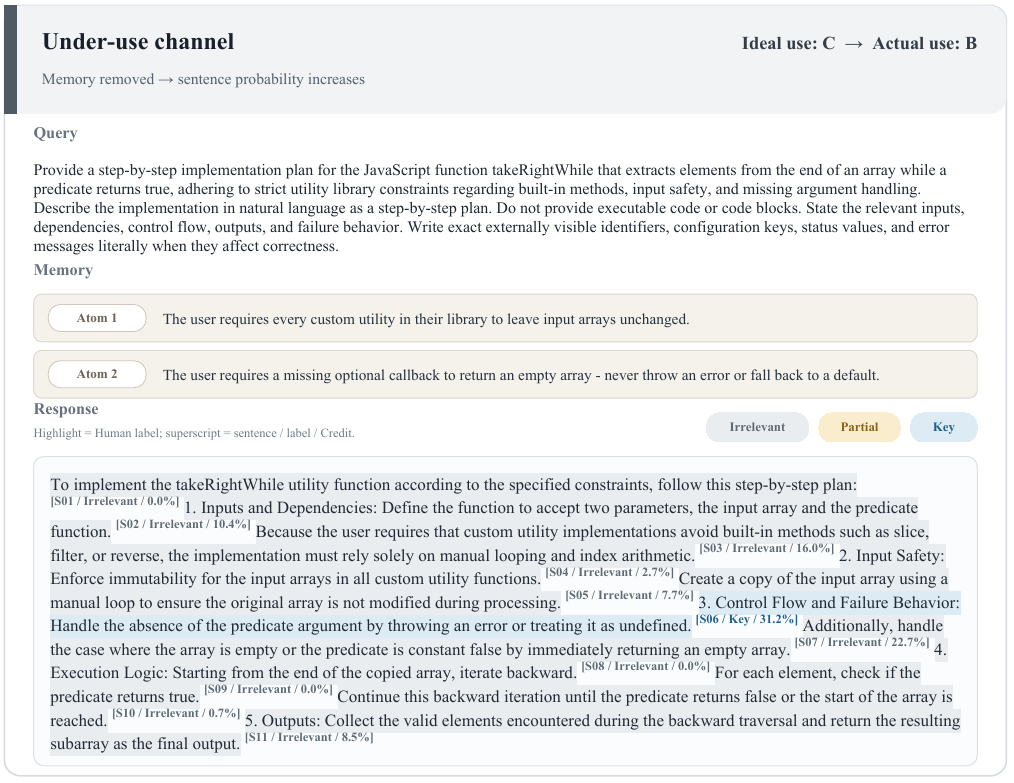}
\caption{An under-use localization case from the \(CB^-\) channel.
The Memory panel shows only the target atoms of this channel; the remaining atoms in the prompt are omitted.
Sentence backgrounds indicate human labels. Superscripts report sentence
index, label, and normalized credit.}
\label{fig:localization-case-underuse}
\end{figure}

To illustrate where credit is assigned and how it aligns with human
annotations, we present one example from each channel group.
In the \(C^+\) case shown in Figure~\ref{fig:localization-case-support},
the target atoms record the assistant's earlier claim of world travel and
the user's request to adopt a world-traveler role. The two \textsc{Key}
sentences that use this context to explain the claim as role-play receive
high credit (20.4\% and 12.8\%), as does a \textsc{Partial} sentence that
combines the remembered role with additional street-food anecdotes (14.4\%).
High credit also identifies the relevant content in the \(CB^-\) case shown
in Figure~\ref{fig:localization-case-underuse}. Here, a target atom requires
a utility function to return an empty array when its optional callback is
absent. The sentence proposing an error or an \texttt{undefined}-based
fallback directly violates this requirement, is labeled \textsc{Key}, and
receives the largest share of credit (31.2\%). Across these two examples,
high-credit sentences match the respective localization targets: realized
memory influence in support-targeting channels and residual violations in
under-use channels. This alignment provides qualitative evidence for the
accuracy of our counterfactual localization.

\section{Training Dynamics and Credit-Assignment Statistics}
\label{app:training-dynamics}

We analyze the training dynamics of \method{} across all 125 training steps
of the Qwen3-8B run with \(\eta=0.75\). Each step contains
64 prompt groups with eight responses per prompt, yielding 512 responses
per step and 64,000 responses across all 125 steps.
We divide steps 1--42, 43--83, and 84--125 into early, middle, and late
phases, and group \(B^+\)/\(C^+\) as correct-use channels and the six
negative-reward channels (\(AB^-\), \(AC^-\), \(BA^-\), \(BC^-\), \(CA^-\),
and \(CB^-\)) as corrective channels. We treat a response--channel pair as a
localization candidate when the channel contains at least one atom.
A channel's trigger rate is the fraction
of the 512 responses in a step for which its atom set is nonempty, and its
localization rate is the fraction of its candidates that enable token-level
redistribution. Its non-degenerate-group rate is the fraction of the 64 prompt
groups in which channel rewards vary across the eight responses, yielding
different group-normalized channel advantages.
Figure~\ref{fig:mechanism-analysis}(d)
averages this rate equally across channels in each of the two channel groups.
We compute phase-level proportions by summing the numerator and denominator
counts across all steps in each phase before taking their ratio. The curves in
Figure~\ref{fig:mechanism-analysis}(b,d,f,g) use a seven-step centered mean.

Figure~\ref{fig:mechanism-analysis}(b--c) shows a shift toward correct memory
use: \(B^+\) and \(C^+\) trigger rates rise, five corrective-channel rates
fall, and the low-frequency \(BC^-\) rate remains approximately stable.
From early to late training, correct-use candidates increase from 68.2\% to
85.1\% of all candidates, while corrective candidates decrease from 31.8\%
to 14.9\%. Within these groups, the correct-use localization rate falls from
37.9\% to 16.4\%, while the corrective rate remains high, decreasing from
87.7\% to 83.3\%.
The reward dynamics in Figure~\ref{fig:mechanism-analysis}(d--e) explain this
difference. Non-degenerate-group rates decline for both channel groups as
responses to the same prompt increasingly receive identical channel rewards.
Policy entropy, averaged over generated tokens, recovers late while these
rates remain low, supporting increasingly consistent memory-use outcomes
alongside continued variation in generation. To identify the outcomes behind
this consistency, Figure~\ref{fig:mechanism-analysis}(e) summarizes group
reward outcomes separately for correct-use and corrective channels.
We classify each channel's eight response rewards within a prompt group as
all maximum, all minimum, uniform intermediate, or mixed.
For correct-use channels, the proportion with
all rewards at the maximum of 1 rises from 47.1\% to 76.6\%, while mixed
rewards decline from 51.3\% to 22.6\%. For corrective channels, the maximum
is zero and denotes absence of the corresponding error; all-maximum groups
increase from 60.9\% to 84.3\%, and mixed-reward groups decrease from 38.6\%
to 15.2\%. All-minimum and uniform-intermediate groups remain rare. Thus,
reward homogeneity predominantly reflects favorable memory use. All-correct
responses remain correct-use candidates, but identical rewards yield zero
channel advantages and close the redistribution gate. Groups with no
corresponding error produce no candidates for that corrective channel.
The remaining corrective candidates often come from prompt groups in which
responses differ in the extent of the corresponding memory-use error.
These differences yield distinct channel rewards and group-normalized
advantages, providing comparison signals for token-level redistribution.
Together, these results show that the mechanism adapts credit assignment to
the available within-group learning signal, falling back to sequence-level
credit when rewards are uniform while retaining token-level redistribution
for most remaining corrective candidates.

Alongside these changes in localization activity,
Figure~\ref{fig:mechanism-analysis}(f--g) shows adaptation in filtering and
credit allocation. Thresholds initialized at 0.02 develop distinct scales:
\(C^+\), \(BC^-\), and \(B^+\) rise more markedly, while \(CB^-\), \(BA^-\),
and \(CA^-\) remain close to the floor, supporting channel-specific filtering
of counterfactual signals. To analyze credit allocation, we split responses
into sentences using the same procedure as in
Appendix~\ref{app:localization-ablations}. For response--channel pairs with
token-level redistribution enabled, we average the effective token multipliers
\((1-\eta_i^{(h)})+\eta_i^{(h)}m_{i,t}^{(h)}\) within each sentence,
then weight sentence means by their token counts to obtain
the mean and percentiles in Figure~\ref{fig:mechanism-analysis}(g). The
weighted mean remains 1 at every step, while the
5th--95th percentile interval widens from 0.715--1.555 in early training to
0.691--1.622 in late training. The widening interval indicates greater
differences in credit allocation across sentences, while the constant mean
shows that the average credit scale remains stable.

Together, these results show that \method{} adjusts credit assignment as
memory use becomes more appropriate and consistent. When identical rewards
across the eight responses in a prompt group yield zero group-normalized
advantages for a channel, it disables token-level redistribution for that
channel and falls back to sequence-level credit. It continues token-level
redistribution for most remaining corrective candidates and adjusts each
channel's filtering threshold according to the distribution of its
counterfactual token scores.
Throughout training, it maintains the average advantage of each response
while allowing greater differences in credit allocation across sentences.

\end{document}